\documentclass[letterpaper]{article}
\usepackage{uai2020}
\usepackage[margin=1in]{geometry}

\usepackage{pdfpages}
\usepackage{cite}
\usepackage{float} 
\usepackage{subfigure} 
\usepackage{graphicx}

\usepackage{multirow}
\usepackage{algpseudocode}
\usepackage{amsmath}
\usepackage{geometry}
\usepackage{algorithmicx}
\usepackage{algpseudocode}
\usepackage{stfloats}
\makeatletter
\newif\if@restonecol
\makeatother

\usepackage[linesnumbered,ruled,lined]{algorithm2e}

\usepackage{times}

\title{Lineage Evolution Reinforcement Learning}

\author{Zeyu Zhang, Guisheng Yin} 

\begin{document}

\maketitle

\begin{abstract}
We propose a general agent population learning system, and on this basis, we propose lineage evolution reinforcement learning algorithm. Lineage evolution reinforcement learning is a kind of derivative algorithm which accords with the general agent population learning system. We take the agents in DQN and its related variants as the basic agents in the population, and add the selection, mutation and crossover modules in the genetic algorithm to the reinforcement learning algorithm. In the process of agent evolution, we refer to the characteristics of natural genetic behavior, add lineage factor to ensure the retention of potential performance of agent, and comprehensively consider the current performance and lineage value when evaluating the performance of agent. Without changing the parameters of the original reinforcement learning algorithm, lineage evolution reinforcement learning can optimize different reinforcement learning algorithms. Our experiments show that the idea of evolution with lineage improves the performance of original reinforcement learning algorithm in some games in Atari 2600.
\end{abstract}

\section{INTRODUCTION}

Deep reinforcement learning has been widely concerned since it was proposed. Deep Q-Networks (Mnih, 2013, 2015) algorithm combines Q-learning with convolutional neural network, introduces the concept of deep reinforcement learning, and achieves excellent results in Atari games (Bellemare, 2013). Double DQN (Hasselt, 2016) solves the overestimate problem in the original DQN by separating the selection function and the evaluation function of the action. Prioritized experience replay (Schaul, 2015) improves the efficiency of sampling by marking the priority of sample data. Dueling DQN (Wang, 2016) separates the value of environment and action, and improves the accuracy of value evaluation of action. A3C (Mnih, 2016) integrates DQN with policy gradient, absorbs the idea of asynchronous and multi-step bootstrap targets (Sutton, 1988), adjusts the balance between variance and deviation, and improves the convergence speed of agent performance. NAF (Shixiang Gu, 2016) enables agents to be used in continuous action space by using positive definite matrix to represent advantage function. Distribution RL (Bellemare, 2017) changes the value evaluation object from average value to distribution, avoiding the evaluation deviation caused by average value fitting. Noisy DQN (Fortunato, 2017) adds noisy linear layer to DQN algorithm to adjust the exploratory strategy so that the agent can explore stably. Rainbow (Hessel, 2017) integrates six DQN extensions, including DDQN, Prioritized replay, Dueling networks, Multi-step learning, Distributional RL and Noisy Nets, which further improves the performance of the agent on Atari 2600, and analyzes the impact of different components on the performance. QR-DQN (Dabney, 2017) added the concept of quantile regression to Distributional RL and proposed. By introducing risk neutral policy, IQN (Dabney, 2018) was proposed based on QR-DQN, and further improves the performance of quantile network.

Genetic algorithm, like reinforcement learning algorithm, is also an algorithm that borrows from natural biological characteristics. Genetic algorithm (Holland, 1975) was first proposed in 1975. This paper systematically expounded the theoretical basis of genetic algorithm. Then, Goldberg further improved the theory of genetic algorithm (Goldberg, 1989), and laid the foundation of modern genetic algorithm research. In recent years, the ideas of genetic algorithm, evolution strategy and neural network are gradually integrated. Kenneth proposed NEAT (Kenneth, 2002, 2009, 2012) by combining genetic algorithm with neural network, and further optimized NEAT in 2009 and 2012.

Referring to the intelligent evolution process of natural organisms, we believe that a single genetic algorithm or reinforcement learning algorithm cannot achieve the most efficient learning process. In order to improve the learning ability of agents, we propose an agent population learning system, which combines the core ideas of neural network, reinforcement learning algorithm and genetic algorithm. We try to integrate some mainstream artificial intelligence algorithms and propose a general agent design system.

According to the agent population learning system, we propose lineage evolution reinforcement learning (LERL). Lineage evolution reinforcement learning algorithm combines the core idea of deep reinforcement learning algorithm and genetic algorithm. In the application of genetic algorithm, we use parameter perturbation and network crossover to replace the original mutation operation and crossover operation. Referring to the lineage factor in nature, we integrate lineage factor and contemporary performance into the agent evaluation module to retain the potential performance value of the agent. The value of lineage will be determined by the historical performance of agents, and will be inherited in the process of evolution.

The idea of lineage evolution can be applied to different reinforcement learning algorithms. In this paper, we choose the reinforcement learning algorithm (DQN, C51, Rainbow and IQN) provided by Dopamine (Castro, 2018) framework as the research object, and use its baseline data as the reference data for experimental analysis. The experimental results show that LERL is superior to the original algorithm in terms of learning speed and performance.

%

\section{BACKGROUND}

This part will introduce the core idea and principle of reinforcement learning and genetic algorithm.

\subsection{REINFORCEMENT LEARNING}

\textbf{Markov Decision Process.} Markov Decision Process (MDP), which is a tuple $(S; A; T; r; \gamma)$, is a kind of mathematical model to simulate agent strategy and reward in the environment. In the tuple, $S$ stands for the state space, $A$ stands for the action space, $T$ is the transition function, $r$ is the reward function, and $\gamma$ is a discount factor. Its theoretical basis is Markov chain. According to the different degree of perception of the environment, there are some variants of Markov decision process, such as partial Markov decision process and constrained Markov decision process. Markov decision process has Markov property, that is, the state of the current moment is only related to the state of the last moment and the action taken at the last moment, but not to the state before the last moment. Its conditional probability formula is
\begin{equation}
p(s_{i+1}|s_i,a_i,...s_0,a_0)=p(s_{i+1}|s_i,a_i).
\end{equation}
\textbf{Deep Q-Networks.} Deep Q-Networks (DQN) is the first algorithm that combines Q learning with neural network. Most of the original Q-learning algorithms access the state and corresponding rewards in the form of tables, which is difficult to deal with the complex environment of state and action. Neural network just makes up for its shortcomings. In DQN, the selection of agent action is decided by neural network. When an agent makes a decision, the current observation of the environment is the input of the neural network, and the Q value of each discrete action is the output of the neural network. The action with the largest Q value is selected to execute. The loss function of training DQN is
\begin{equation}
(R_{t+1}+\gamma_{t+1}{\rm{max}}q_{\bar{\theta}}(S_{t+1},a')-q_{\theta}(S_t,A_t))^2.
\end{equation}
\textbf{Distributional RL.} Distributional RL is a kind of reinforcement learning algorithm which can fit the distribution of action value. In the original dqn, only one value is output for each action, which represents the mean value of the action. Other attributes of the action value distribution are not considered too much. However, some values with large distribution gap may have the same mean value, which makes a lot of important information lost in value function fitting. In order to solve this problem, the output of distribution RL is no longer the traditional mean value, but the value distribution of actions. After setting the upper and lower limits of all action values, the network will output the value distribution of an action. In their paper, the most representative algorithm is C51, which outputs a 51 dimensional vector for each action, and each element represents the probability of its corresponding value.

\textbf{Rainbow.} In Rainbow (Dopamine), researchers implemented the three most important modules in the original Rainbow algorithm: n-step Berman update, priority experience replay and distributional RL. Among them, prioritized replay means that the algorithm defines the priority of training data according to absolute TD error. The algorithm is more inclined to take out the samples with high priority when sampling, which greatly improves the efficiency of agent learning.

\textbf{Implicit Quantile Networks.} Based on C51, Dabney proposed QR-DQN. QR-DQN takes quantile as fitting object, which improves the accuracy and robustness of value distribution evaluation. After that, Dabney added an extra neural network to the original QR-DQN to dynamically determine the fitting accuracy, and proposed Implicit Quantile Networks (IQN), which has risk sensitive property.

\subsection{GENETIC ALGORITHM}

The idea of genetic algorithm comes from Darwin's law of natural selection. Genetic algorithm encodes the solution of the problem first, and then optimizes the solution of binary encoding format through selection, mutation and crossover. Genetic algorithm keeps the individuals with high fitness through the fitness evaluation function, and takes the retained individuals as the parents of the new population to generate new solutions. Genetic algorithm mainly includes five parts: fitness evaluation, selection operation, mutation operation, crossover operation and judgment of termination conditions.

\textbf{Fitness evaluation.} Fitness evaluation is the basis of selection, crossover and mutation in genetic algorithm, which is used to measure the performance of an individual in the current generation. Generally, the fitness function needs to be adjusted according to the application environment.

\textbf{Selection operation.} After getting the fitness evaluation results of all individuals in the population, genetic algorithm will select the individuals with higher fitness from the current population according to certain rules as the parent samples of the next generation population. The purpose of this operation is to retain genes with high fitness in the evolution process.

\textbf{Mutation operation.} The mutation operation originates from the mutation of biological chromosomes in nature. In mutation operation, genetic algorithm determines whether to perform mutation operation according to mutation probability, and randomly selects one or a group of basic bits for mutation.

\textbf{Crossover operation.} The object of traditional crossover operation is two string binary codes. Following the crossover and recombination of biological chromosomes, two binary codes are used to realize crossover operation.

\textbf{Judgment of termination conditions.} The goal of genetic algorithm is to find the local optimal solution rather than the global optimal solution in the high-dimensional solution space. Therefore, it is difficult for genetic algorithm to stop by finding the global optimal solution. There are two common termination conditions. The first is to limit the number of evolutions. The second is to terminate the algorithm when the performance reaches a certain score or the performance fluctuation is small.

\section{AGENT POPULATION LEARNING SYSTEM}

\begin{figure*}[htbp] 
\centering 
\includegraphics[width=0.95\textwidth]{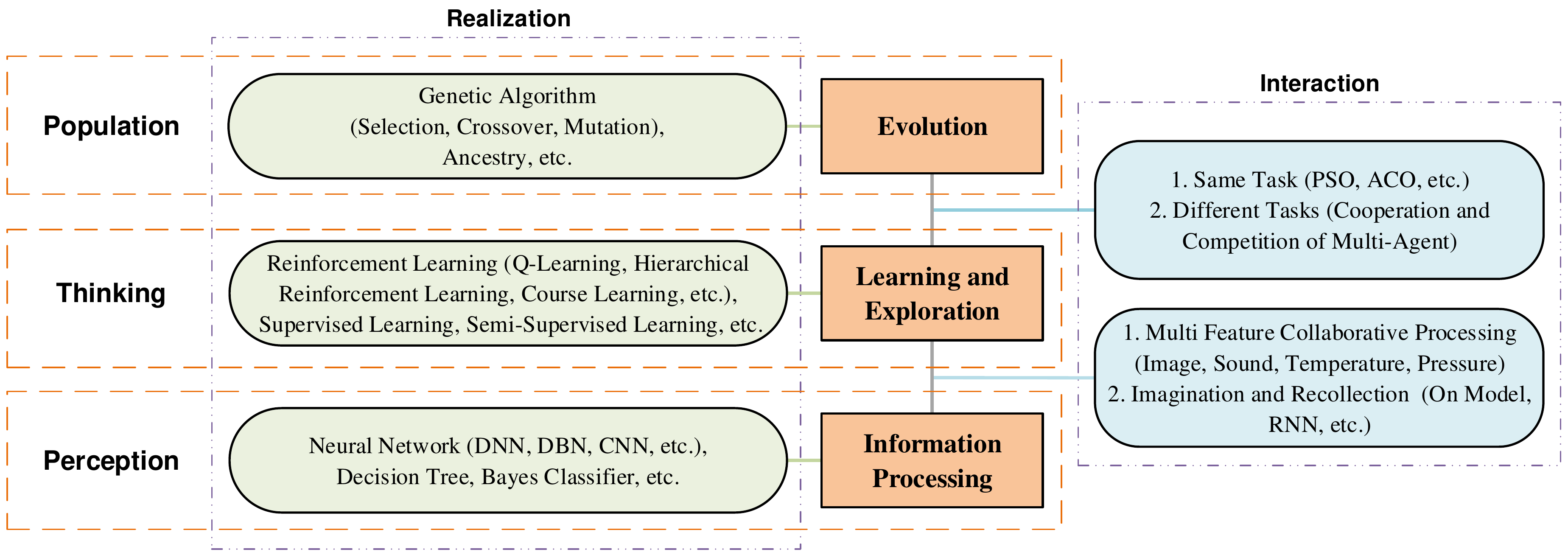} 
\caption{General Agent Population Learning System} 
\end{figure*}

At present, a lot of artificial intelligence algorithms are based on the specific problems and analyze the characteristics of the problems. For example, natural language processing focuses more on the structure of language, and image recognition focuses more on the processing of image features. If we analyze these tasks from a more abstract perspective, we can get a more general algorithm design pattern. On the one hand, a general algorithm system can reduce the limitations of algorithm application, on the other hand, it can improve the efficiency of algorithm design when researchers deal with new tasks. The inspiration of agent population learning system proposed in this paper comes from human perception, learning and evolution. In the general agent population learning system (GAPLS), we divide the architecture of agent into three levels, from bottom to top, they are perception layer, thinking layer and population layer.

In the perception layer, the agents mainly complete the task of information processing. The agents use some algorithms directly deal with the external information obtained by a certain perception medium, abstract this information, and transmit the information obtained from the abstraction to the thinking level. The function of perception layer corresponds to the function of perception organ in human intelligence. The machine learning of this module focuses on improving the recognition ability of agents to environmental information, aiming to extract higher-level features through complex and large-scale representation information.

The task of the thinking layer is to study and explore. In many current algorithms, the concept of thinking is usually blurred or bound with the perception layer. In fact, there is a certain deviation from the way of thinking of human intelligence. Although human intelligence is essentially the transmission of neurotransmitters, perception and decision-making are realized by different brain regions. Therefore, we think that the thinking layer should be treated as a single layer. At present, many algorithms can be regarded as thinking layer algorithms. The input of the thinking layer is a feature group transmitted from the perception layer, and its output is the decision-making of the agent. The task goal of the thinking layer is to gradually improve the adaptability of agents to the environment without changing the input characteristics, and improve their performance in a specific environment or task through existing data. The exploration of agents in the thinking layer is relatively stable.

The population layer is at the top of the whole agent population learning system, and it is also the layer with the most exploratory and the least call times in the learning process of agent population. The population layer of agents corresponds to the population of human beings, both of which take independent intelligent units as basic computing units. The most typical algorithm in the population layer of the same task is genetic algorithm, and the algorithms of different tasks are mainly focused on the field of game theory. The goal of population level is to break the limitation of current learning, jump out of local optimal solution, and try to make breakthroughs to improve the adaptability of agents to the environment. In addition, if the population layer adopts the elite strategy, the excellent and rapid evolution of a single agent can improve the average performance of the whole agent population in a limited time, and drive other agents to make leaping progress.

Task objectives at different layers are relatively independent. However, due to the difference of perception information, task characteristics and the definition of environment, we still need to supplement the interaction modules between different layers to improve the whole agent population learning system and ensure its universality. The interaction between population layer and thinking layer can be divided into the same task and different tasks according to different task settings. To some extent, interaction patterns will affect the way of evolution.

In the interaction process of thinking layer and perception layer, if the perception mode is not unique, the interaction process may need to preprocess the feature groups acquired by different perceptions to ensure the unity of dimensions. In addition, some tasks involving prediction or timing can be supplemented with imagination and recall modules in the interaction process to improve the pertinence of tasks.

According to the design of agent population learning system, we give the corresponding algorithm flow. In Algorithm 1, we define the structure of perceptions, agents and population.In the evolution operation, we keep the evaluation, selection, mutation and crossover in genetic algorithm, and adopt the elite strategy in the agent selection. In order to ensure the accuracy of the evaluation, we add a potential performance factor in the evaluation process. Potential performance can be derived directly from past performance or predicted based on the mapping relationship between past environment and past performance, and updated in each generation.

\begin{algorithm}[htbp]
  \caption{General Agent Population Learning}
  \KwIn{Data set $D$ and task set $T=\{t_0,...,t_{N_{task}}\}$}
  \KwOut{Agent population $E$}
  Initialize perceptions $P(x;\omega)$ \\
  Initialize agents $A(P_0,...,P_{N_{perception}};\theta)$ \\
  Initialize agent populations $E_t=\{A^{t}_0,...,A^{t}_{N_{agent}}\}$ of the same task \\
  Initialize agent population $E=\{E_0,...,E_{N_{task}}\}$ \\
  \For{\rm generation $g \in \{0,...,N_{generation}\}$}
  {
    \If{$g>0$}
    {
      \For{\rm task $t \in T$}
      {
        \For{\rm agent $A^{t}_{k} \in E_t$}
        {
          Compute performance ${\rho}^{t,k}_g$\\
          ${\Gamma}^{t,k}_g \gets w_{\rho}{\rho}^{t,k}_g+w_{\phi}{\phi}^{t,k}_g$  \\
        }
        Divide $E_t$ into $E^{1}_t$, $E^{2}_t$ and $E^{3}_t$ \\
        Generate $E^{c}_t$ and $E^{m}_t$ from $E^{1}_t$\\
        $E_t \gets E^{1}_t \cup E^{2}_t \cup E^{c}_t \cup E^{m}_t$ \\
        \For{\rm agent $A^{t}_{k} \in E_t$}
        {
          Update potential ${\phi}^{t,k}_g$  \\
        }
      }
    }
    \For{\rm iteration $i_a \in \{1,...,I_a\}$}
    {
      \For{\rm agent $A' \in E_0 \cup ... \cup E_{N_{task}}$}
      {
        \For{\rm $p \in \{0,...,N_{perception}\}$}
        {
          \For{\rm iteration $i_p \in \{1,...,I_p\}$}
          {
            Update ${\omega}_{A',p}$ by ML algorithm \\
          }

        }
        Update ${\theta}_{A'}$ by ML algorithm \\
      }
      \If{\rm environment is interactive}
      {
        Interact with the environment \\
        Update $D$ \\
      }
    }
  }
\end{algorithm}

In the general agent population learning algorithm, we train the mapping function of thinking layer and perception layer independently according to the structure of agent population learning system. Taking the neural network algorithm as an example, we divide the agent network into two parts. The perception part can generate as many different features as possible through semi supervised learning, while the thinking part uses another completely independent network for supervised learning. When the network of the thinking part is trained, the network parameters of the perception part remain unchanged. When the agent interacts with the environment, we make the thinking part and the perception part work together. This training mode has the following advantages: 1) The perception function can be reused. 2) In line with the design concept of agent population learning system. 3) Training is more efficient.

In the general agent population learning algorithm, if ${i_p}$ is equal to 1, the agent's thinking layer and perception layer can be regarded as bundling training, which is consistent with the current DQN and its variant algorithms.Considering the different types of tasks, we take environment interaction as an option in the learning process. In addition, under the condition of ensuring that the agents of evolution operation are in the same generation, agents with the same task can save running time in parallel when they are learning and updating training data.

\section{LINEAGE EVOLUTION REINFORCEMENT LEARNING}

According to the general agent population learning system and the corresponding algorithm proposed above, we bring the current reinforcement learning algorithm into it and propose lineage evolution reinforcement learning algorithm (LERL). LERL is a derivative of the above algorithm in a specific environment. In LERL, we train the thinking and the perception of the agent together. In addition to the combination of reinforcement learning algorithm and genetic algorithm, LERL also introduces lineage factor to evaluate the potential performance of agents.

\subsection{EVALUATION}

The evaluation of agent performance is the first step of agent evolution. The evaluation process occurs before each selection, mutation and crossover, and after the learning of all agents of the previous generation. Agents perform evolution operations on a regular basis in the process of reinforcement learning. When the agent population is in the evolution generation, we evaluate the agent performance $\Gamma$. Because of the parallelism of the algorithm, the time for each agent to reach the evolutionary operation is different, so we also set the evolution lock for the agent population. When an agent reaches the evolutionary operation, the evolution lock is closed. When all agents are ready for evolution operation, the evolution lock is opened, and the algorithm evaluates the performance of all agents.

Due to the randomness and complexity of the environment, only using the current performance of agents for evaluation will not be able to accurately measure the performance of agents in the following situations:

1) The current performance of agent is accidental, which is far from the result of re running in the same environment. Due to the uncertainty of the environment, the performance of the current agent may be accidental, and the agent cannot guarantee that the performance of the next run is the same as the current one.

2) Although the current performance of the agent may be good, its mobility is poor. Reinforcement learning is a kind of unsupervised learning algorithm, but due to the limitation of learning samples, its learning process will also have the phenomenon of over fitting. When over fitting occurs, the agent can perform well in the same or similar environment as the current learning samples. However, the agent may not perform well in other environments under the same settings, its learning potential is low.

3) There is a big gap between the current learning samples and the historical samples. Although the current performance of the agent is not good, it has great potential. As we all know, with the running of reinforcement learning algorithm, the replay buffer used for agent training will be constantly updated. Although the agent will follow certain random principle to extract samples during training, it is undeniable that the learning samples in different periods have different characteristics. In the learning process, some agents with good performance may encounter an environment having a large gap with the samples in the current buffer. In this case, the current performance of these agents may be lower than that of history or other agents of the same generation. However, this kind of agent may have high potential. After training in the current unfamiliar environment, this kind of agent will have stronger robustness.

In order to solve the above problems, based on the general agent population learning system, we imitated the lineage theory in human reproduction, and added the lineage factor as one of the evaluation indexes. The new evaluation method comprehensively considers the current performance and implicit performance of the agent, thus improving the accuracy of agent evaluation.

The evaluation of agent can be divided into two sub steps. The first step is to normalize the current performance of the agent. The current performance $\rho$ of an agent is represented by the score of the agent in the environment, so the performance distribution in different environments is different. During the comprehensive evaluation, it is necessary to ensure that the value range of lineage is consistent with that of current performance. After normalizing the current performance of the agent, the performance and lineage have the same value range [0,1]. The second step is computing the comprehensive evaluation value $\Gamma$ of the agent according to the lineage value $\phi$ and the current performance $\rho$. The weight of lineage value and current performance is $w_{\phi}$ and $w_{\rho}$ respectively.

\subsection{LINEAGE VALUE UPDATING}

After the evaluation of the agent, the algorithm enters the lineage value updating phase. In LERL, lineage will be updated with the learning of agents, and the new lineage value will be determined by the historical lineage value and the current performance of agents. The process of lineage value renewal can be divided into three steps. The first step is to give the lineage update value $\Delta\phi$ according to the current performance of the agents. The second step is to generate new lineage value according to the old lineage value and lineage renewal value. The third step is to normalize the new lineage value.

When calculating the lineage update value, we rank the agents according to their current performance. The higher the performance of the agent, the higher the lineage update value. The maximum lineage update value is 1, and all of them are positive.
\begin{equation}
\Delta\phi_i=\frac{A_n-A^{rank}_{i}+1}{A_n}
\end{equation}
\begin{equation}
\phi_i=\frac{(\Delta\phi_i+\phi_i*\zeta_o)-{\rm min}(\Delta\phi_i+\phi_i*\zeta_o)}{{\rm max}(\Delta\phi_i+\phi_i*\zeta_o)-{\rm min}(\Delta\phi_i+\phi_i*\zeta_o)}
\end{equation}
It should be noted that the renewal of lineage value must be strictly carried out after the comprehensive evaluation. Because if the lineage value update is performed first, the comprehensive evaluation value will consider the current performance twice. Although from the perspective of algorithm execution, only adjusting the weight of performance can achieve the same effect as the current algorithm flow, but multiple consideration of performance will increase the complexity of lineage update function and increase the difficulty of parameter adjustment.

\subsection{SELECTION}

In the selection operation part, we adopt the elite strategy to select the agents to be retained. According to the comprehensive evaluation value of agents, the agent population can be divided into three parts.

The first part is the elite agents $E_1$. Each time a selection operation is performed, several agents with the best performance are selected as elite agents. As the parents of new agents, elite agents will generate new agents through mutation or crossover operation.The second part is general agents $E_2$. The performance evaluation of general agent is in the middle of the whole population. In the evolution, the part of agent is directly a member of the next generation population.The third part is to eliminate the agents $E_m$ and $E_c$. The performance of the eliminated agents is at the end of the current population, which will be discarded in evolution due to their poor performance. Because the population size is fixed, the positions of the eliminated agents will be replaced by the new agents. The new agents come from the mutation operation and crossover operation of elite agents.

The elite strategy has the following functions: 1) It ensures that the excellent agents will not be lost due to the interference of mutation or crossover. 2) It ensures that the upper performance limit of LERL is not lower than that of the original algorithm. 3) If evolution operation cannot improve the performance of the agent, the whole learning process of the agent will not be affected. 4) If the performance of the agent declines due to the uncertainty of reinforcement learning, the corresponding agent can be discarded in time.

\subsection{MUTATION}

Mutation operation and cross operation are the key to speed up learning and improve performance. Mutation operation can be divided into three parts: agent replication, network mutation and lineage inheritance. Agent replication refers to the online network replication of an elite agent. It should be noted that in the agent replication phase, the mutation operation only deals with the network parameters, and does not copy the data in the replay buffer.

In agent mutation, we disturb the parameters of the network, and the disturbance range is all the layers of online network. The disturbance operation includes two parameters: the first is the probability of disturbance $v_{part}$, and the second is the amplitude of disturbance $v_{range}$. When the network is disturbed, we traverse the network layer by layer, and judge whether the corresponding layer is disturbed according to the disturbance probability. When a layer of the network is disturbed, the shape of the layer network is obtained first, and the disturbed network is generated according to its shape and disturbance amplitude. The parameters in the disturbed network are uniformly distributed.
\begin{equation}
V_{ij}\sim U(1-v_{range},1+v_{range})
\end{equation}
After the disturbed network is generated, we get a new network by Hadamard product between the online parameter matrix and the disturbed parameter matrix.

We introduce crossover operation $O_m$, which is defined as follows:
\begin{equation}
O_m(Q(x,a;\theta)):=Q(x,a;\theta*V)
\end{equation}
The disturbance amplitude $v_{range}$ is very important for mutation operation. If $v_{range}$ is too large, it will lead to the loss of previous learning. If $v_{range}$ is too small, the exploration of mutation operation cannot be guaranteed. In addition, because of the less randomness of the agent in the learning process and the gradual convergence of the algorithm, $v_{range}$ should have different values in different periods. In LERL, with the increase of the number of iterations, $v_{range}$ will decrease.
\begin{equation}
v_{range}={\rm min}(v_{range0}-G_n*v_{range\_decay},v_{range\_min})
\end{equation}
When the disturbance amplitude decays to a certain extent, the disturbance parameters will not change, so as to ensure that the mutation operation remains exploratory.
After the network disturbance, the new mutation agent needs to inherit the lineage value of the original agent. Due to the influence of mutation operation, the inherited lineage value should be attenuated, and its attenuation parameter is $\zeta_m$.

\subsection{CROSSOVER}

The first step in crossover operation is similar to mutation operation, which is to select and copy agents from elite agents as the parent agents in crossover operation. Then, we carry out network crossover operation on the selected two replicated elite agents.

No matter the binary code crossover operation in the traditional genetic algorithm or the expected value crossover operation in the evolution strategy, it is undoubtedly difficult to be directly used in the neural network. Referring to the description of human neural in cognitive neuroscience (Gazzaniga, 2000), we think that convolution network is more emphasis on feature extraction, while full connection layer is more emphasis on decision-making according to the extracted features. The former corresponds to the perception layer in the agent population learning system, and the latter to the thinking layer. Therefore, we split the convolution network and the full connection network in the online network as two basic units of crossover operation. In the crossover operation, we crossover the convolution network part and the full connection network part of the two elite agents, and take the two new networks as the online networks of the new agents.

We introduce crossover operation $O_c$, which is defined as follows:
\begin{equation}
Q(x,a):=Q^f(Q^c(x;\theta^c),a;\theta^f)
\end{equation}
\begin{equation}
O_c(Q_j,Q_k):=Q^f_j(Q^c_k(x;\theta^c_k),a;\theta^f_j)
\end{equation}
Similar to mutation operation, crossover operation also needs to inherit and decay the lineage value of the parent agents, and its decay parameter is $\zeta_c$.

The lineage decay formula of crossover operation is different from that of mutation operation. There are two parent agents of crossover operation, so it is necessary to calculate the average lineage value of parent agents before inheriting and decaying. In addition, because mutation and crossover do not affect each other and there is no strict logical order, the two operations can run in parallel.

\begin{algorithm}[htbp]
  \caption{Evolution with Lineage}
  \KwIn{Agent population $E=\{A_1,...,A_n\}$, Current performance $P=\{\rho_1,...,\rho_n\}$
  and Lineage evaluation $\Phi=\{\phi_1,...,\phi_n\}$}
  \KwOut{Agent population $E$}
  \For{\rm agent $A_i \in E$}
  {
    ${\rho}_i \gets {({\rho}_i-{\rm min(\rho)})}/{({\rm max}(\rho)-{\rm min}(\rho))}$ \\
    Compute comprehensive evaluation value
    ${\Gamma}_i \gets w_{\rho}{\rho}_i + w_{\phi}{\phi}_i$  \\
  }
  Update $\Phi$ using Equation (4)\\
  Divide $E$ into $E_1$, $E_2$, $E_m$ and $E_c$ \\
  \For{\rm agent $A_u \in E_m$}
  {
    Select $A_j$ from $E_1$ \\
    $Q_u \gets O_m(Q_j)$ \\
    $\phi_u \gets \zeta_m\phi_j$ \\
  }
  \For{\rm agent $A_w \in E_c$}
  {
    Select $A^1_k$ and $A^2_k$ from $E_1$ \\
    $Q_w \gets O_c(Q^1_k,Q^2_k)$ \\
    $\phi_w \gets \zeta_c(\phi^1_k+\phi^2_k)/2$ \\
  }
  $E \gets E_1 \cup E_2 \cup E_m \cup E_c$ \\
\end{algorithm}

\section{EXPERIMENT}

We use the data and algorithm in the Dopamine framework for experimental analysis. Among them, the reinforcement learning algorithm for LERL includes DQN, C51, Rainbow and IQN in Dopamine. The running environment includes Asterix, Assault, ChopperCommand and KungFuMaster. The baseline data of each algorithm is provided by Dopamine project.

\begin{figure*}[t]
\centering
\subfigure[Asterix]{
\begin{minipage}[b]{0.23\textwidth}
\includegraphics[width=1\textwidth]{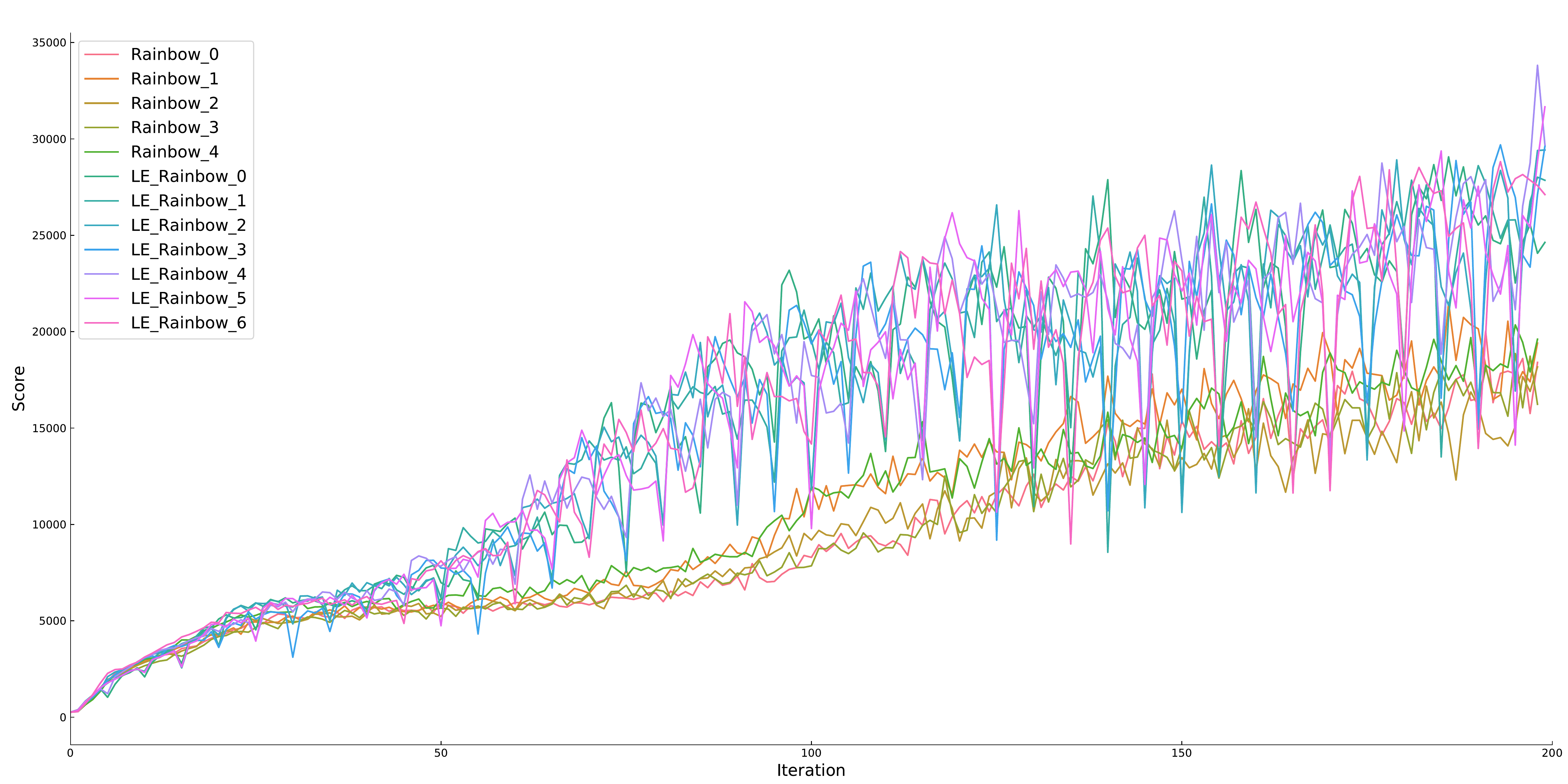} \\
\includegraphics[width=1\textwidth]{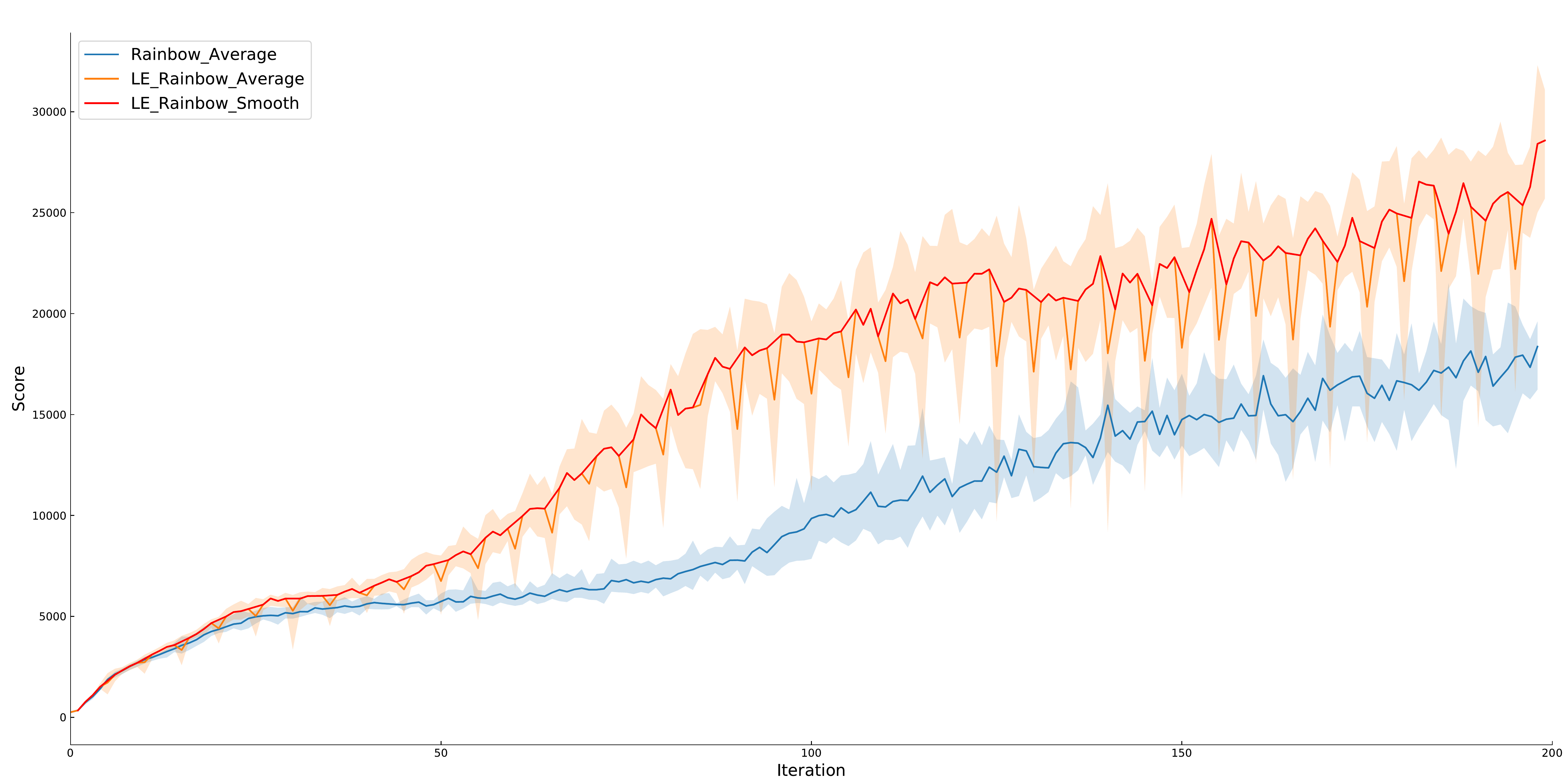} \\
\includegraphics[width=1\textwidth]{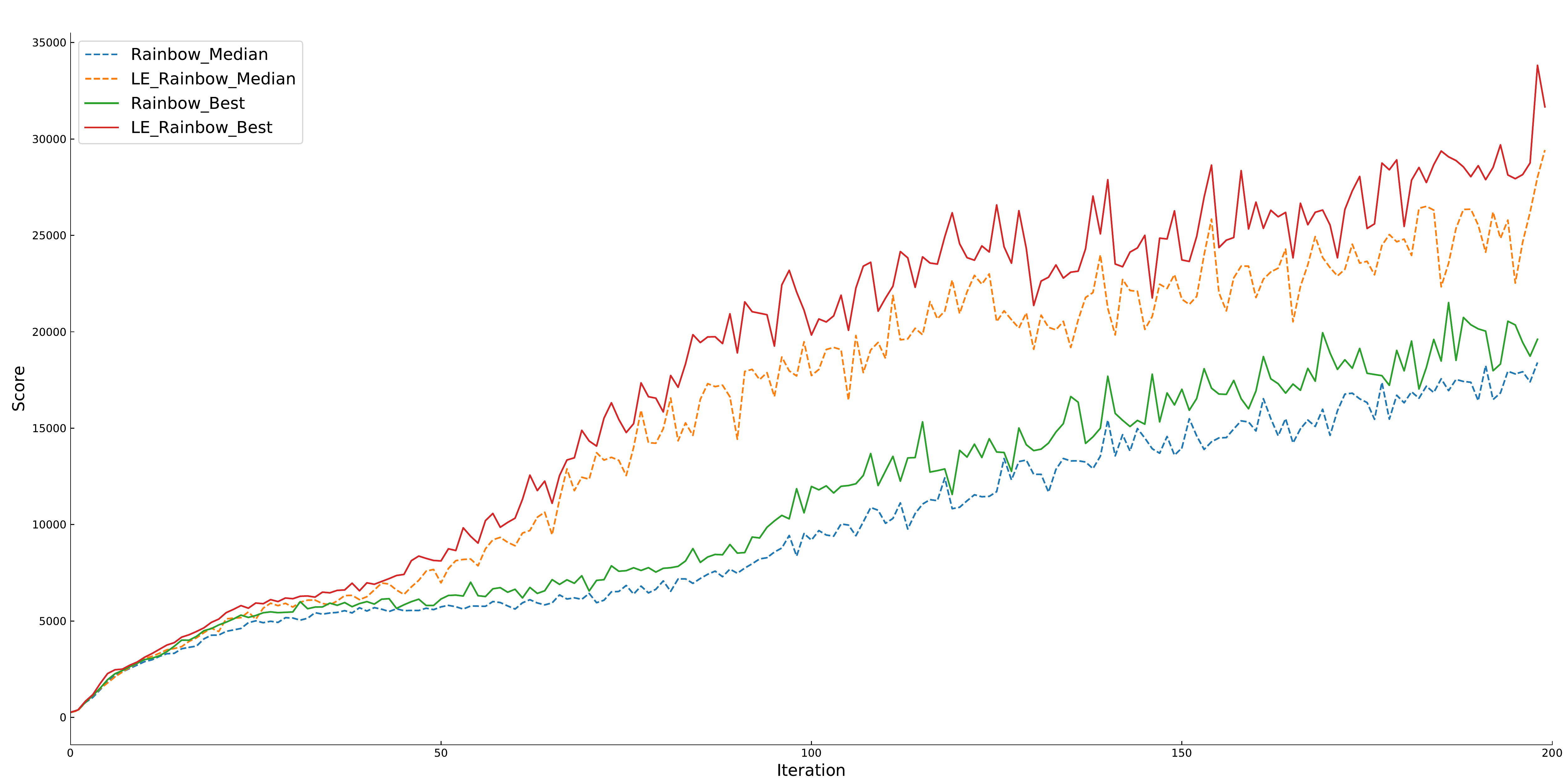}
\end{minipage}
}
\subfigure[Assault]{
\begin{minipage}[b]{0.23\textwidth}
\includegraphics[width=1\textwidth]{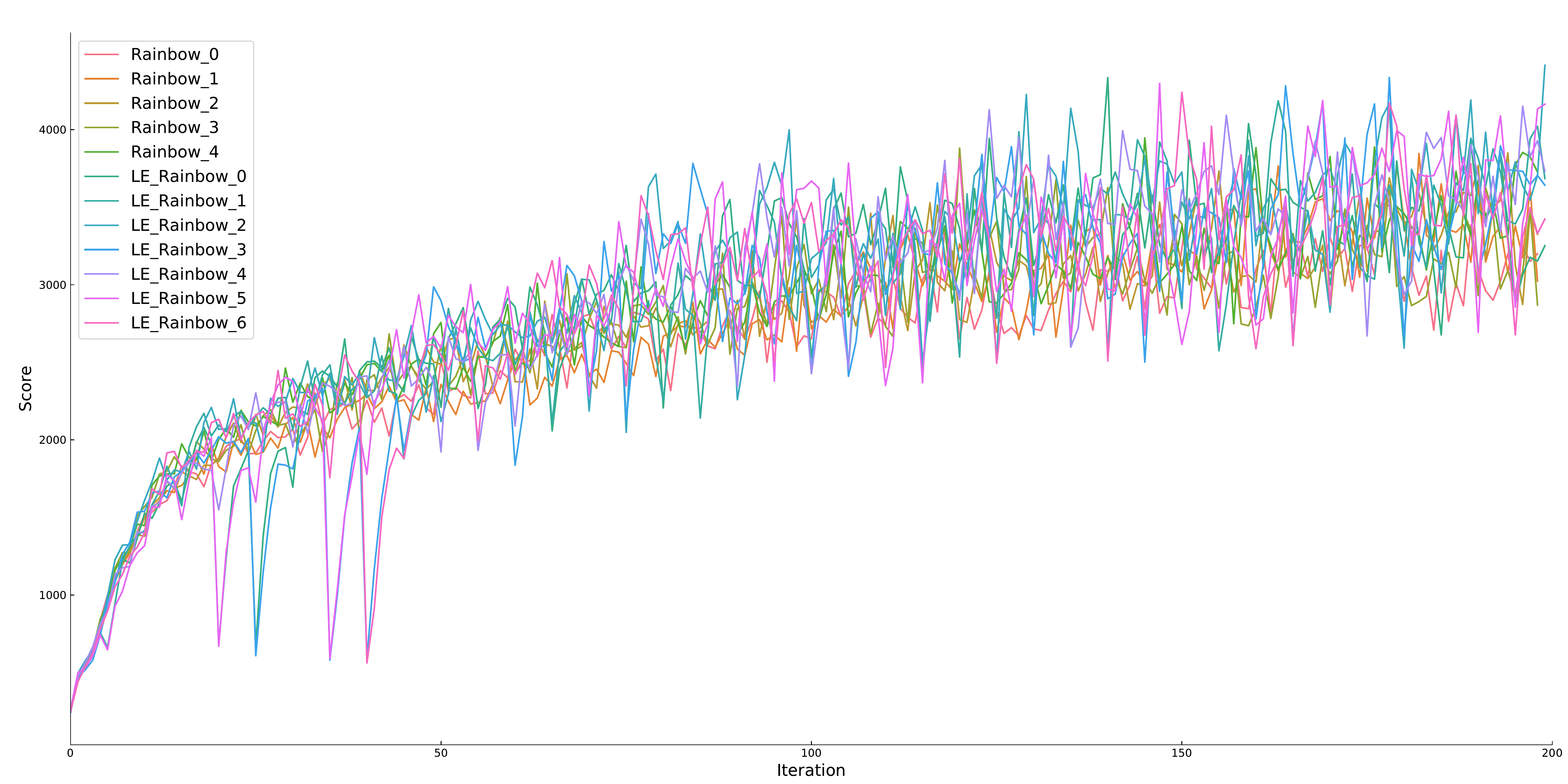} \\
\includegraphics[width=1\textwidth]{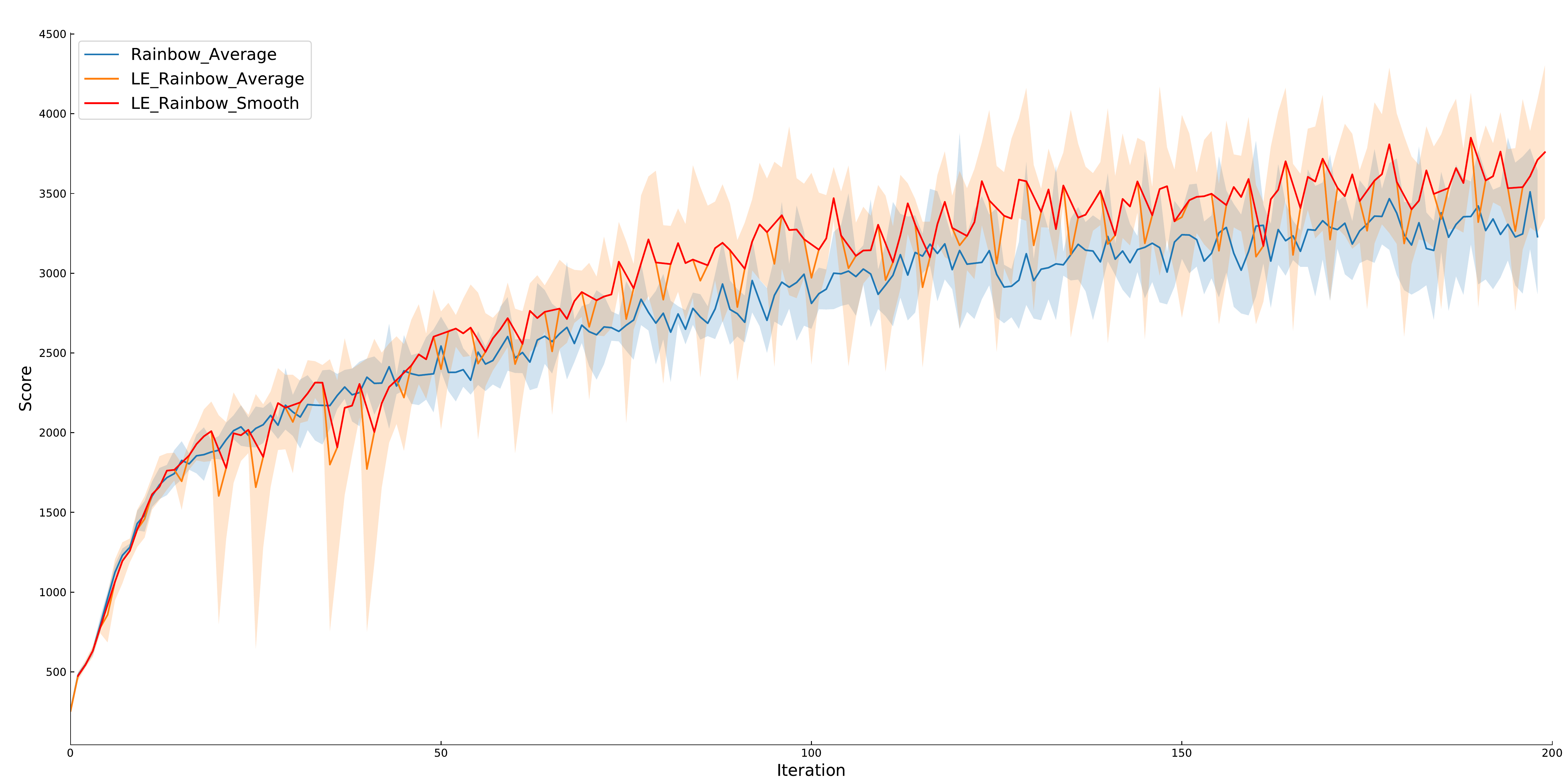} \\
\includegraphics[width=1\textwidth]{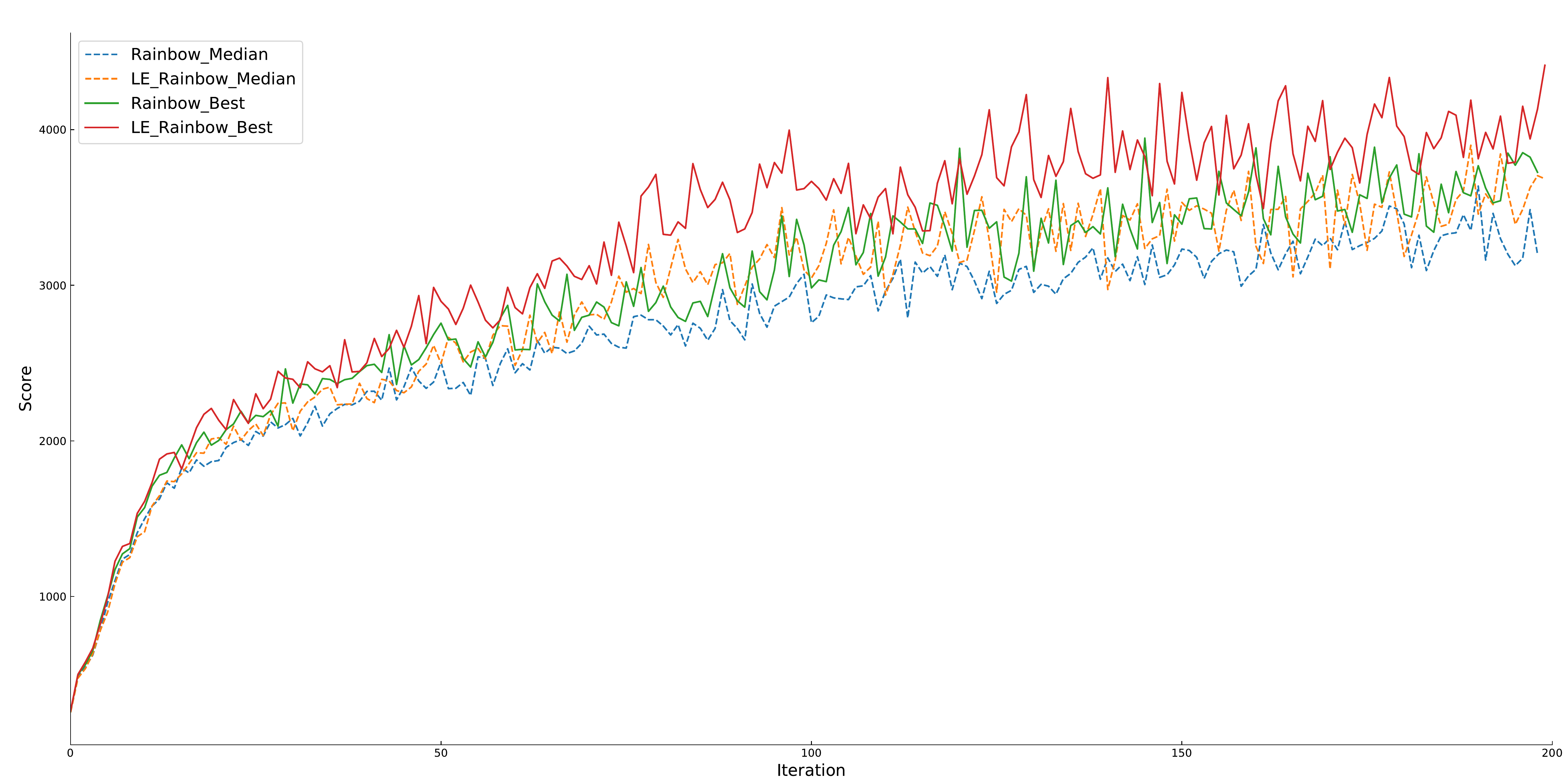}
\end{minipage}
}
\subfigure[ChopperCommand]{
\begin{minipage}[b]{0.23\textwidth}
\includegraphics[width=1\textwidth]{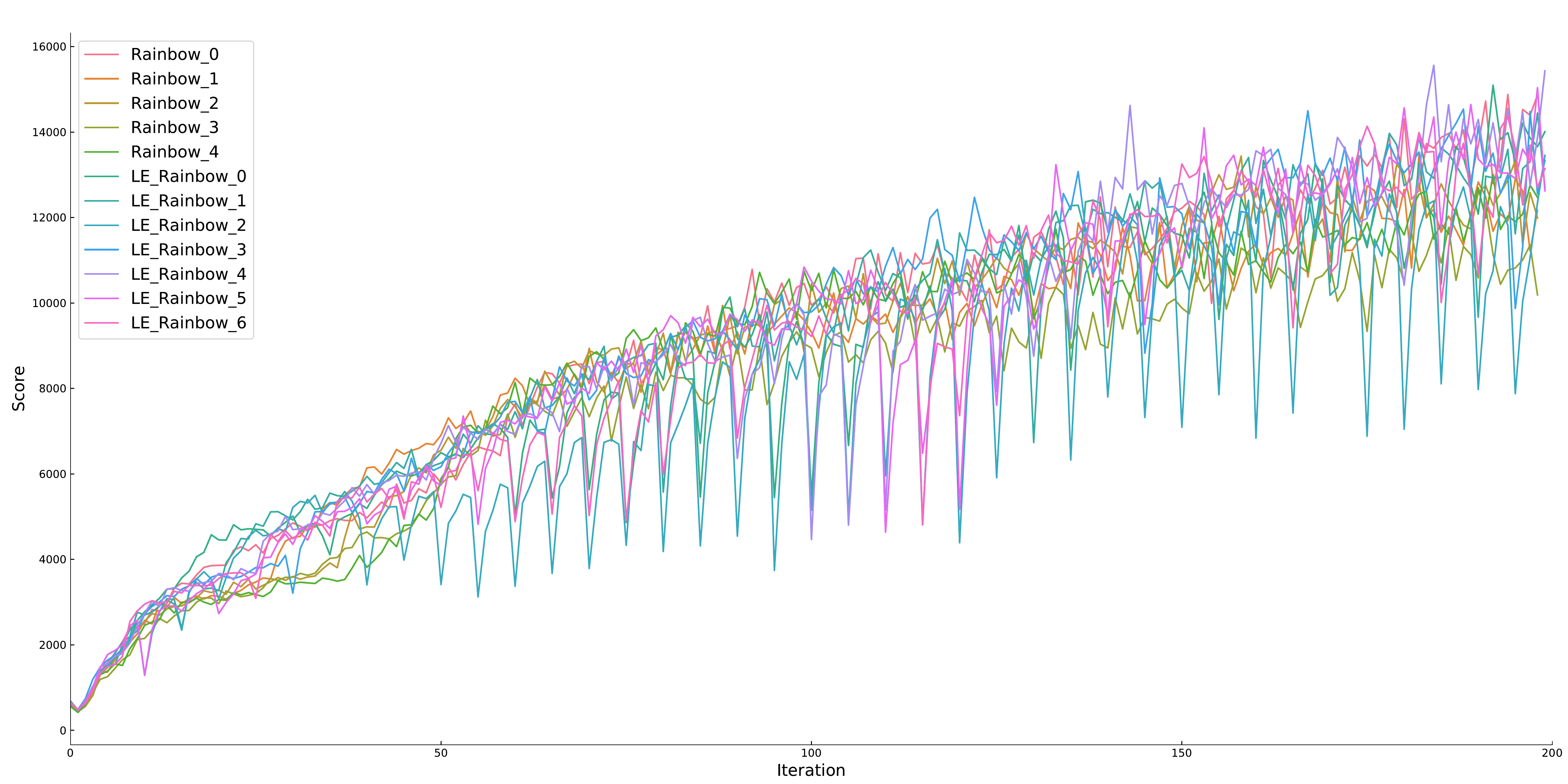} \\
\includegraphics[width=1\textwidth]{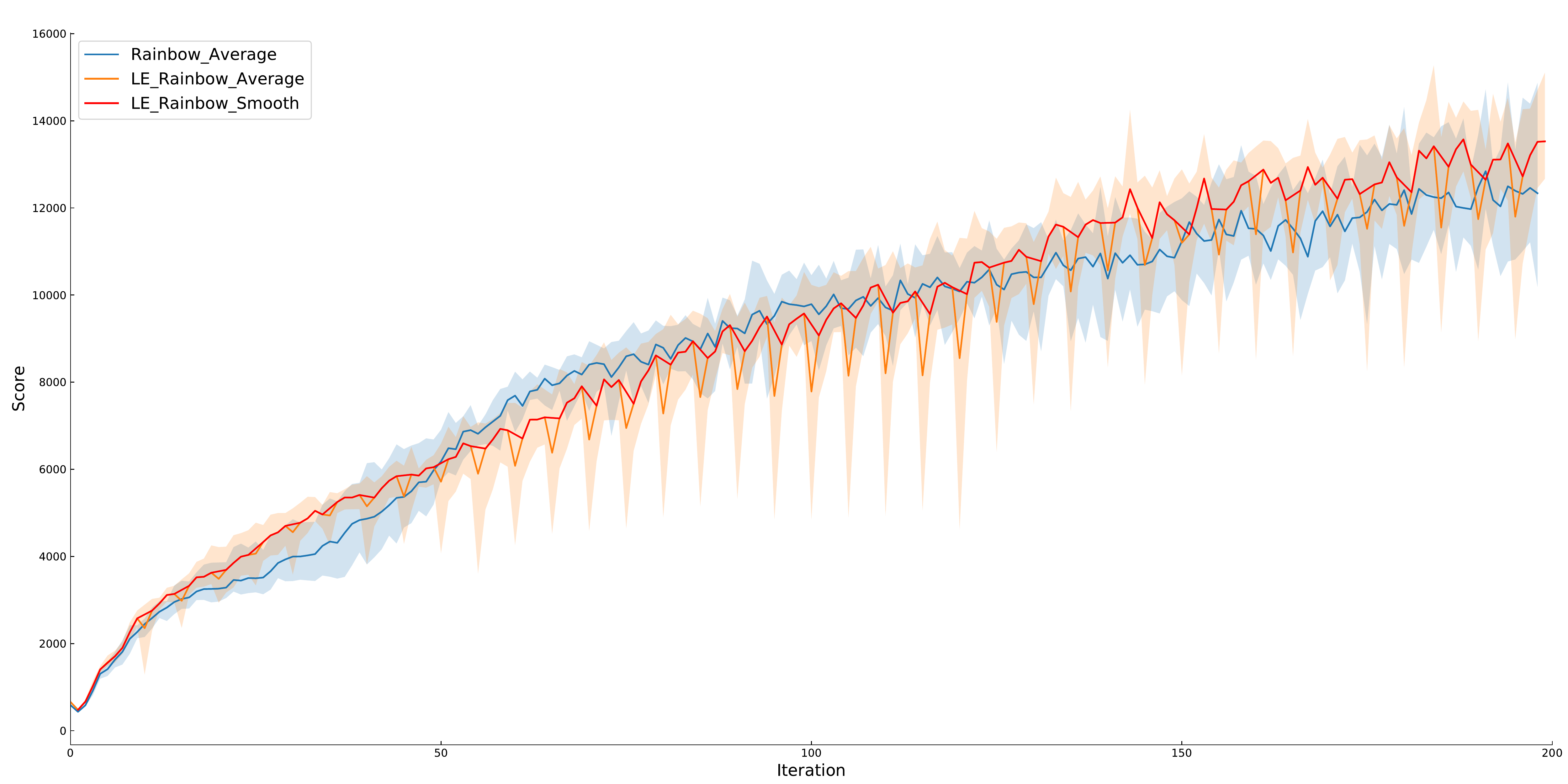} \\
\includegraphics[width=1\textwidth]{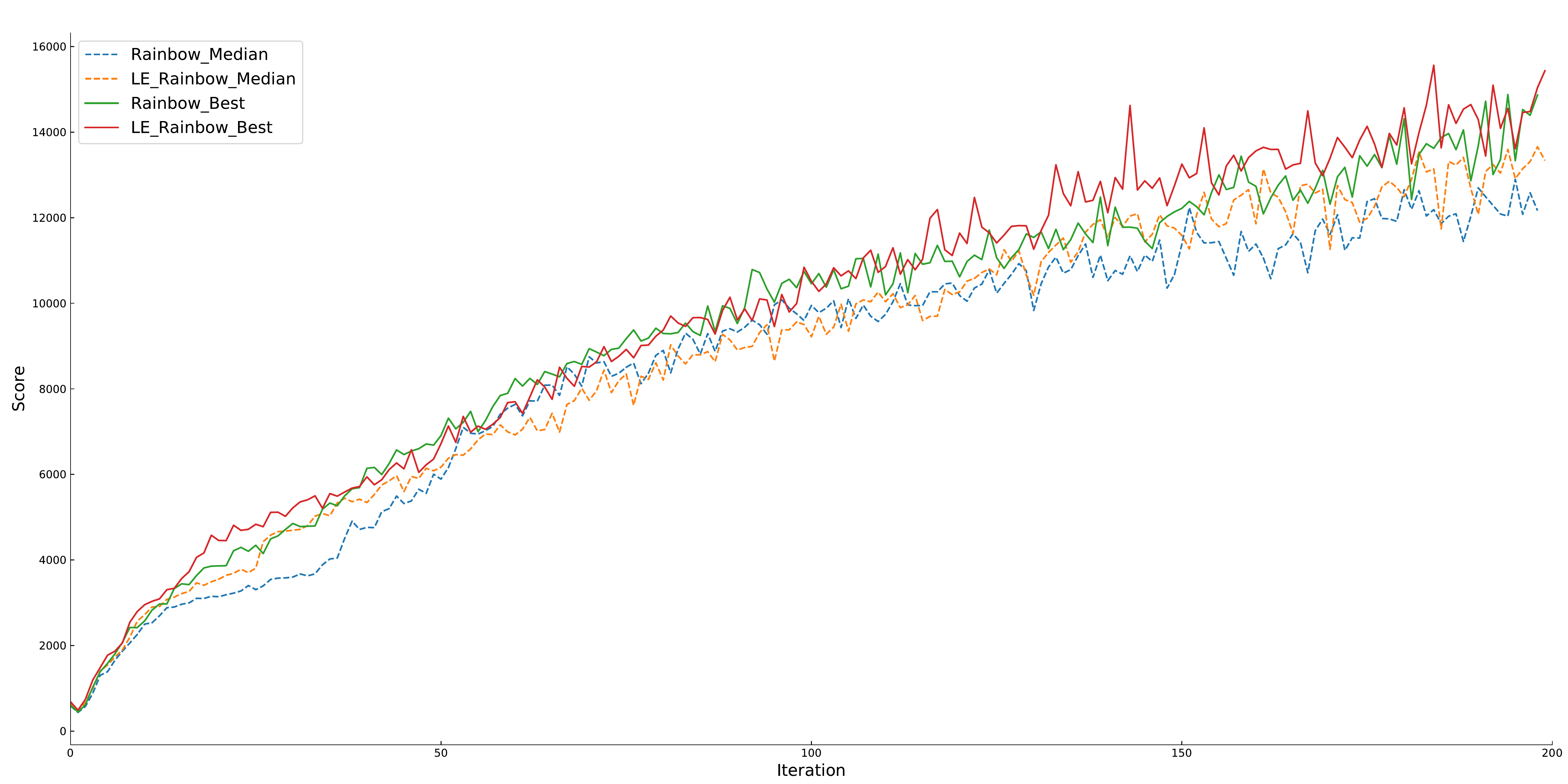}
\end{minipage}
}
\subfigure[KungFuMaster]{
\begin{minipage}[b]{0.23\textwidth}
\includegraphics[width=1\textwidth]{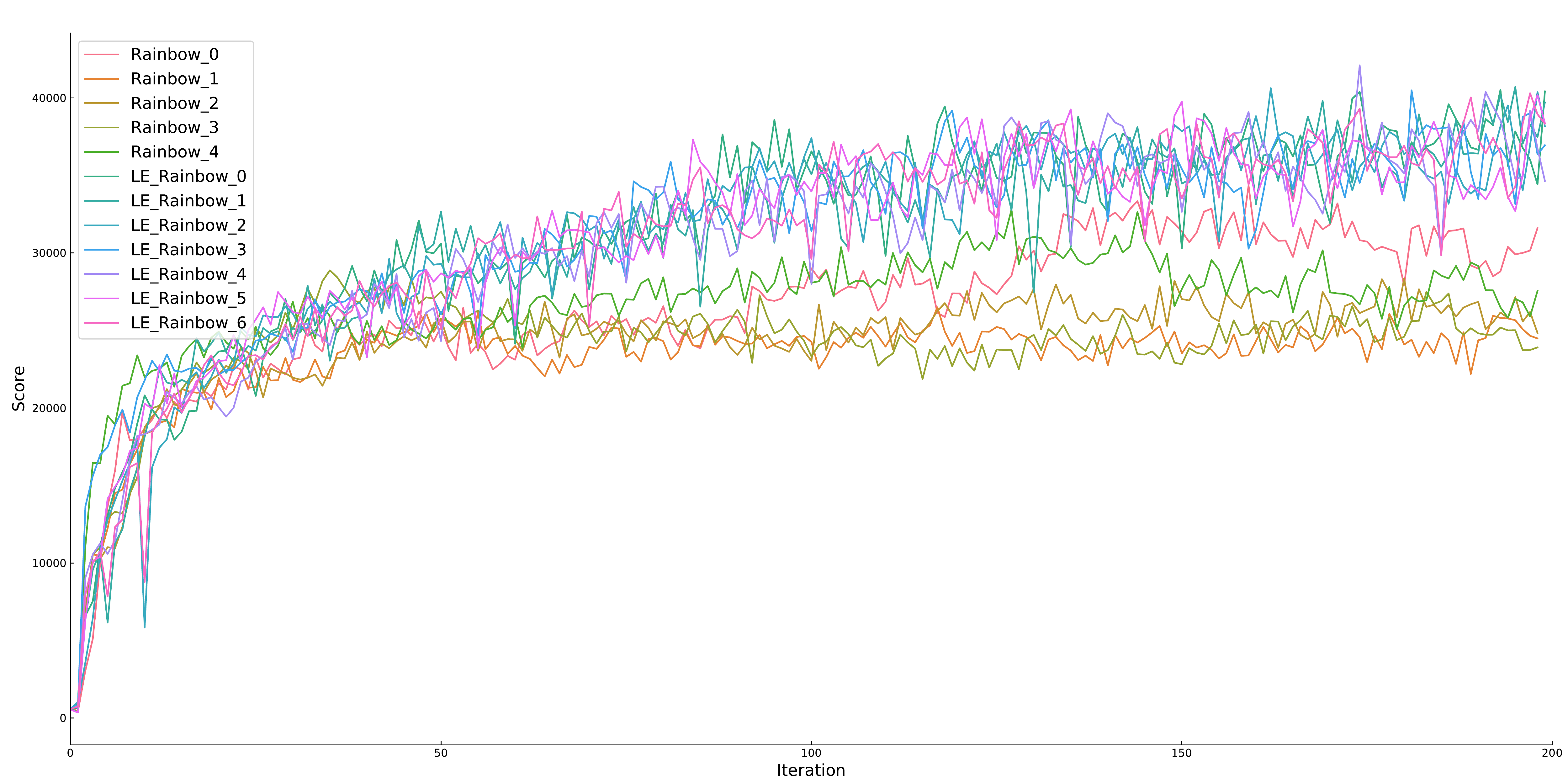} \\
\includegraphics[width=1\textwidth]{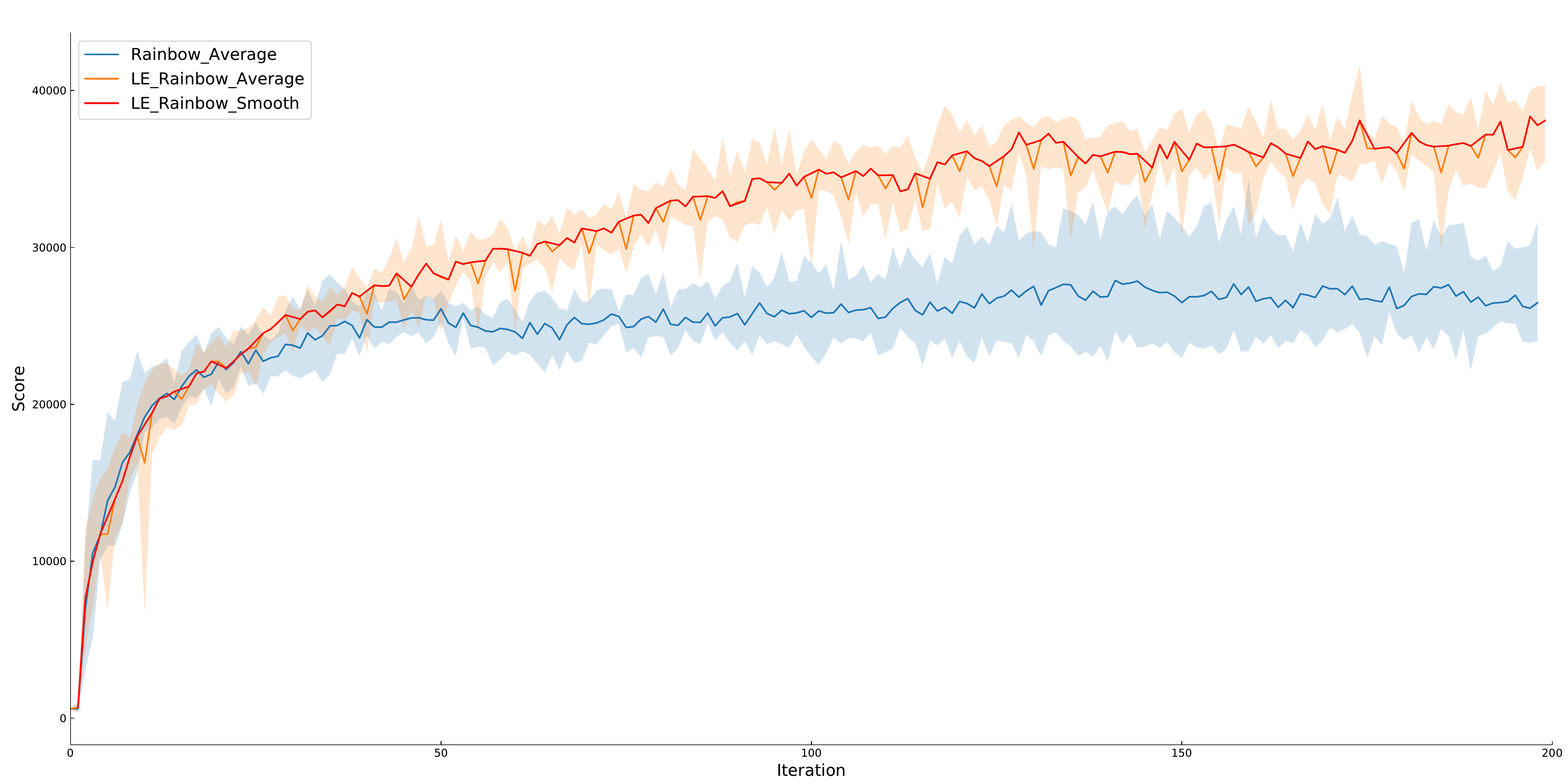} \\
\includegraphics[width=1\textwidth]{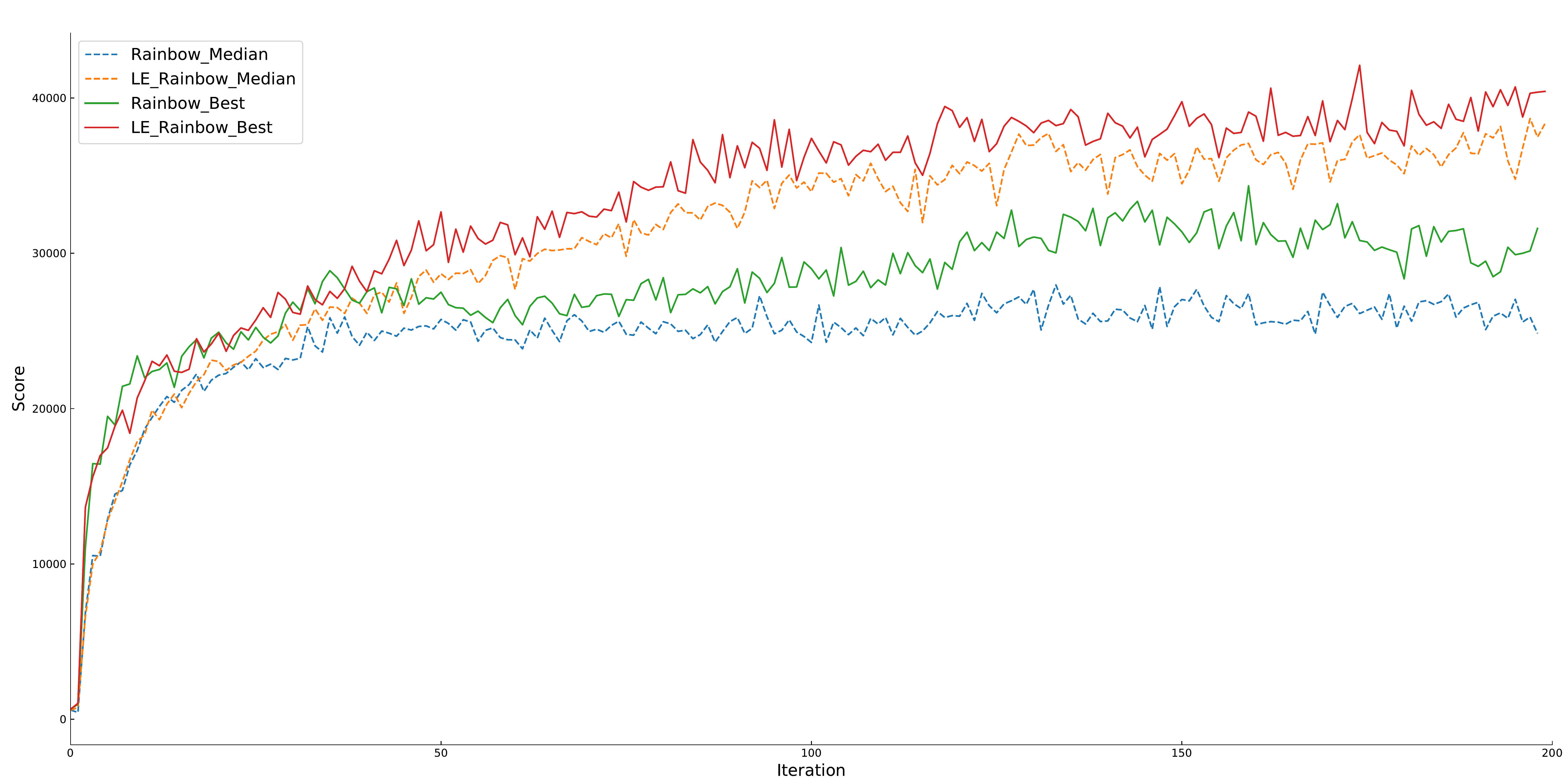}
\end{minipage}
}
\caption{Performance of LE\_Rainbow and Rainbow in Different Games}
\end{figure*}

Figure 2 shows the performance of LE\_Rainbow, which uses Rainbow algorithm as the basic reinforcement learning algorithm, in four games. The images in the first row show the original performance of the agents in Rainbow and LE\_Rainbow. Among them, the agent population of LE\_Rainbow includes 7 agents. In evolution operations, the number of elite agents is 2, the number of general agents is 2, and the number of eliminated agents is 3. Rainbow's curve is the performance data of agents running independently five times. The evolution cycle of LE\_Rainbow is 5 iterations. The images in the second row are the average performance curve of the two algorithms. In addition, due to the randomness of evolution, the performance of LE\_Rainbow algorithm changes greatly, so its average performance is not representative. We introduce the smooth average performance curve LE\_Rainbow\_Smooth for LE\_Rainbow. The images in the third row show the best performance curve and median performance curve of the two algorithms. The best performance curve of the agent represents the best performance of the agent in each iteration, which can better help us to analyze the performance of LE\_Rainbow.

As can be seen from Figure 2, LE\_Rainbow improves the performance of Rainbow greatly in Asterix and KungFuMaster. Whether it's average performance, best performance or median performance, LE\_Rainbow has always been higher than Rainbow since the early stage. In Assault and ChopperCommand, although the performance of LE\_Rainbow is not significantly improved, it also ensures that the performance is not lower than that of Rainbow. In addition, from the best performance curve, we can see that LE\_Rainbow is more exploratory. In the process of exploration, the best performance LE\_Rainbow has achieved is higher than Rainbow.

\begin{figure}[htbp]
\centering

\subfigure[Asterix]{
\begin{minipage}[b]{0.23\textwidth}
\centering
\includegraphics[width=1\textwidth]{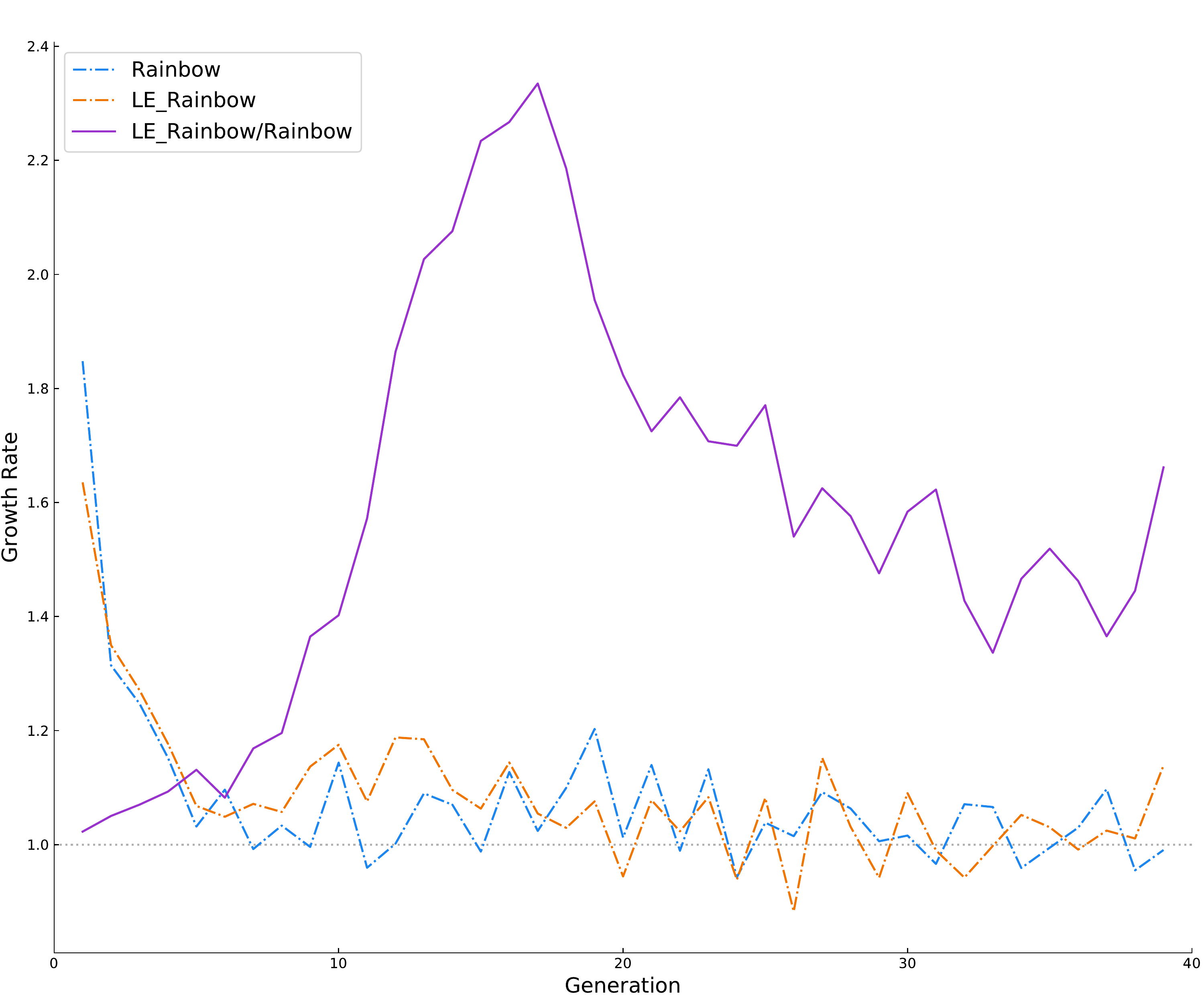}
\end{minipage}%
}%
\subfigure[Assault]{
\begin{minipage}[b]{0.23\textwidth}
\centering
\includegraphics[width=1\textwidth]{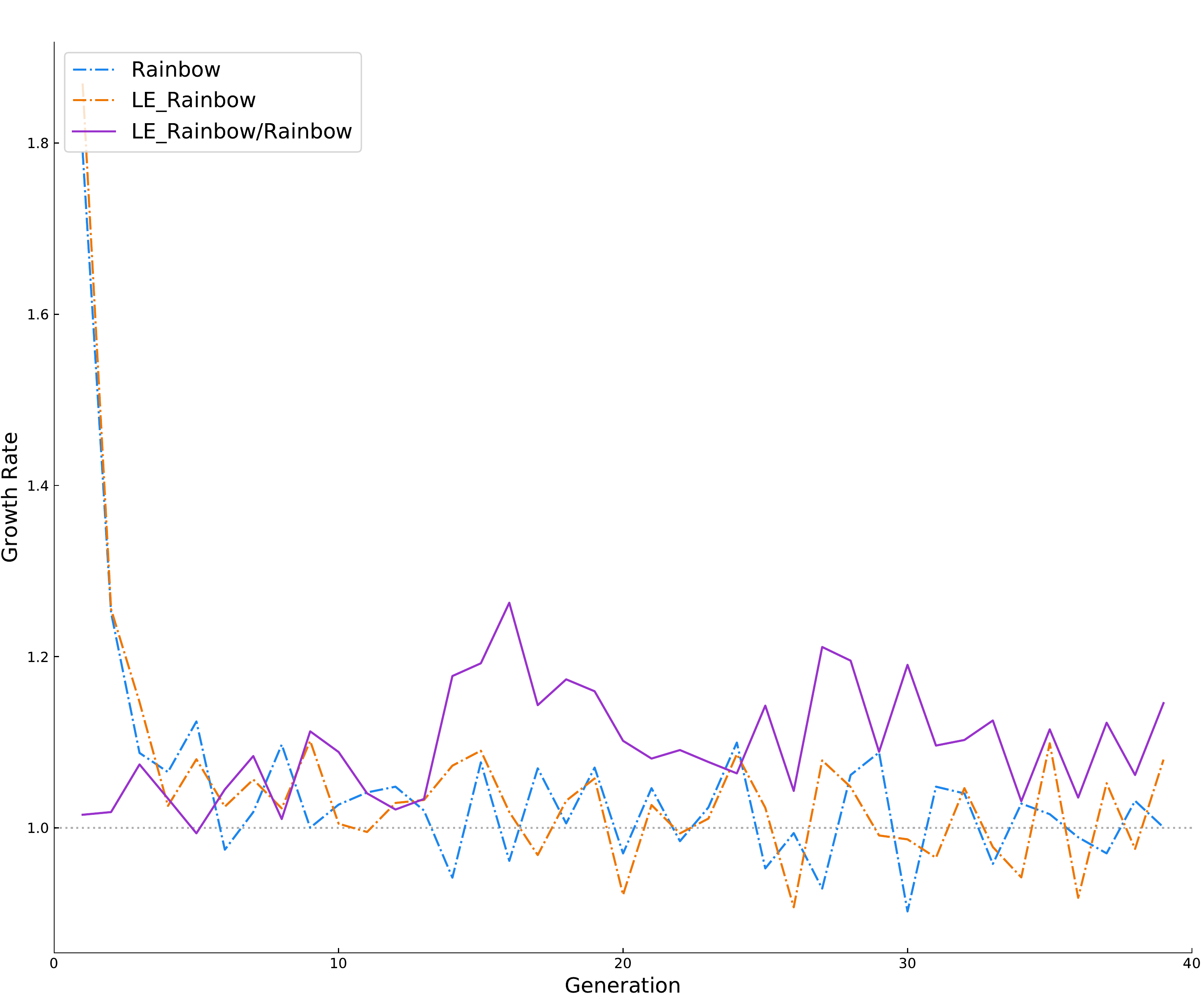}
\end{minipage}%
}%

\subfigure[ChopperCommand]{
\begin{minipage}[b]{0.23\textwidth}
\centering
\includegraphics[width=1\textwidth]{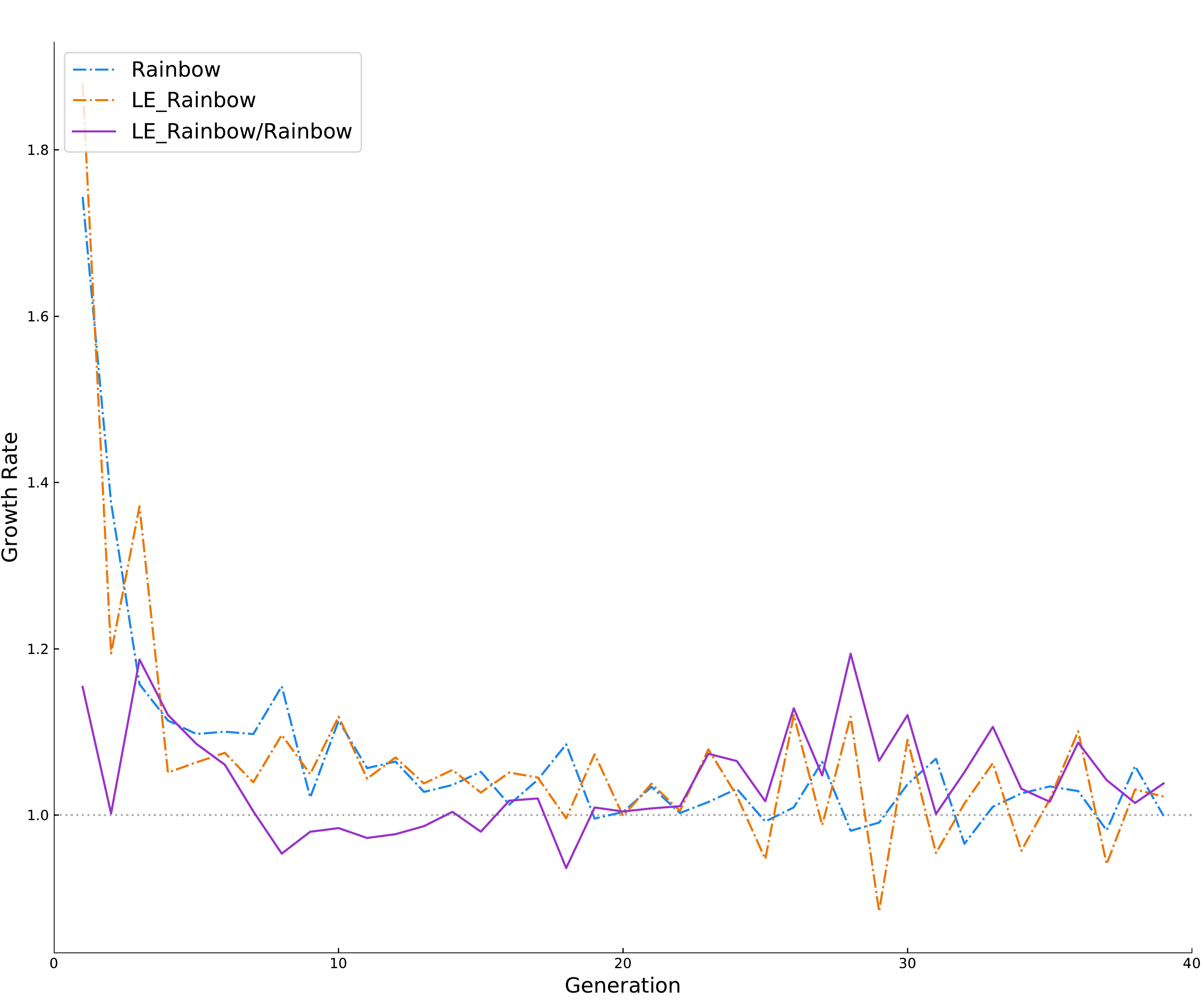}
\end{minipage}
}%
\subfigure[KungFuMaster]{
\begin{minipage}[b]{0.23\textwidth}
\centering
\includegraphics[width=1\textwidth]{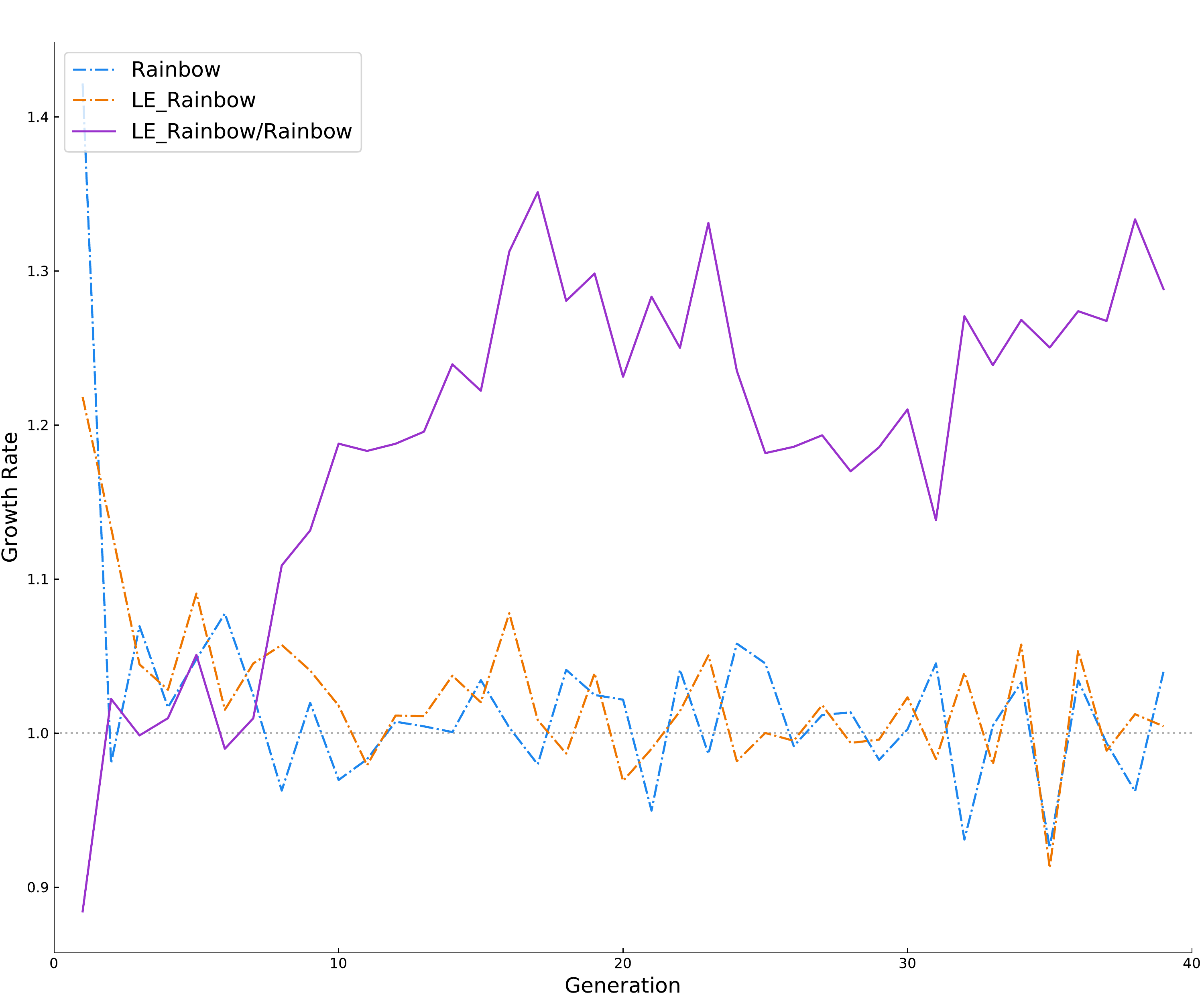}
\end{minipage}
}%

\centering
\caption{The Performance Growth Rate of LE\_Rainbow and Rainbow in Different Games}
\end{figure}

Figure 3 shows the performance growth rate curves of LE\_Rainbow and Rainbow in different evolution cycles and the performance growth rate curves of LE\_Rainbow compared with Rainbow. The performance growth rate is equal to the ratio of the best performance in the current evolution cycle to the best performance in the previous evolution cycle. Two hundred iterations can be divided into 40 evolutionary cycles, that is, 40 generations. The purple curve in Figure 3 represents the ratio of LE\_Rainbow performance to Rainbow performance. The curve is affected by the performance growth rate of the two algorithms.

It can be seen from Figure 3 that the ratio of the best performance of the two algorithms may rise first and then decline during agents evolution. The ascending part indicates that the evolution operation improves the performance of the original reinforcement learning algorithm. The decreasing part can be regarded as a performance convergence phenomenon. The convergence phenomenon is influenced by performance bottleneck (the upper limit of scores available in the environment), function bottleneck (the capacity limit of neural network) and parameter setting (the range of exploration).

\begin{figure}[htbp]
\centering

\subfigure[LE\_DQN]{
\begin{minipage}[b]{0.23\textwidth}
\centering
\includegraphics[width=1\textwidth]{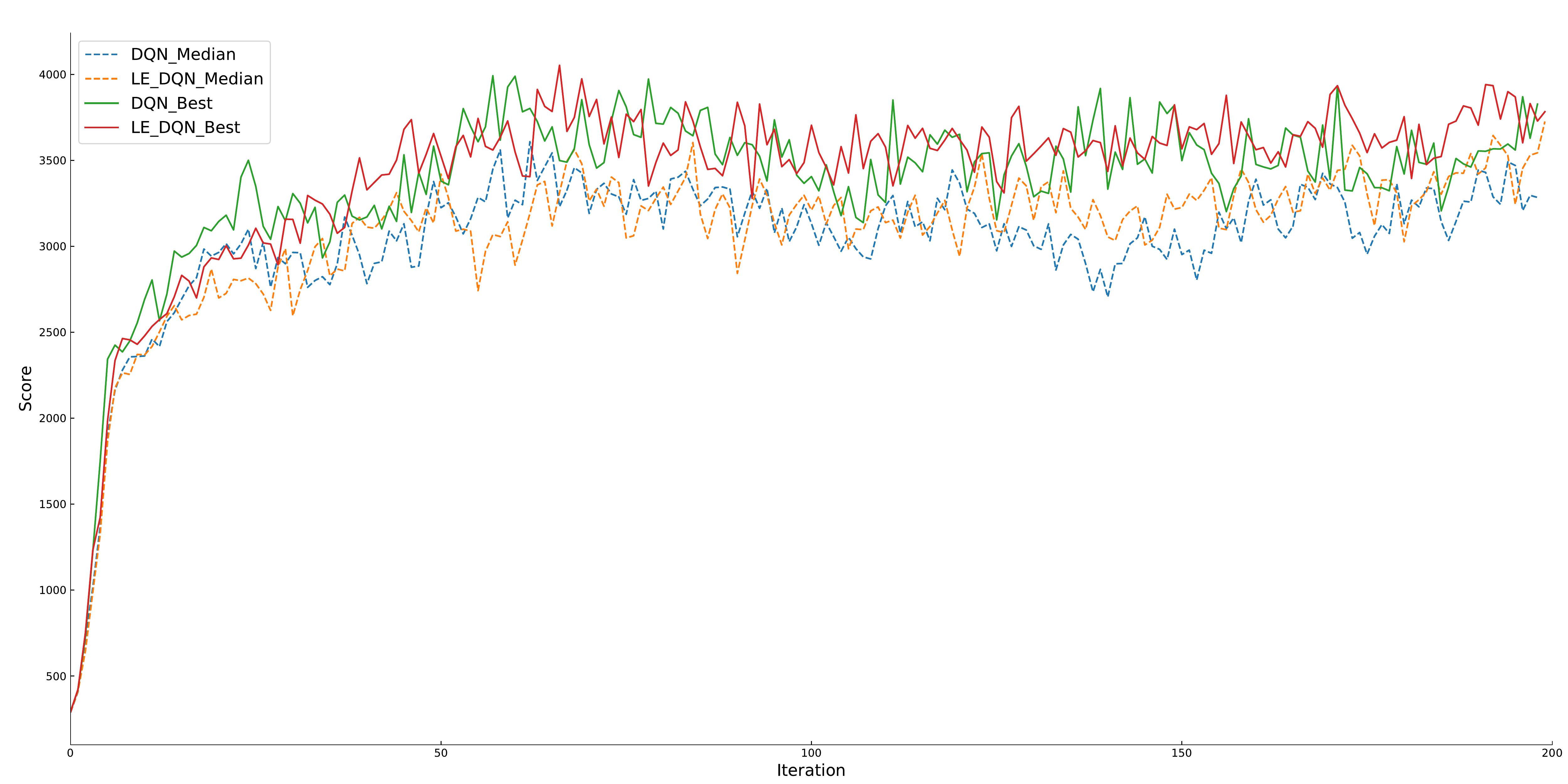}
\end{minipage}%
}%
\subfigure[LE\_C51]{
\begin{minipage}[b]{0.23\textwidth}
\centering
\includegraphics[width=1\textwidth]{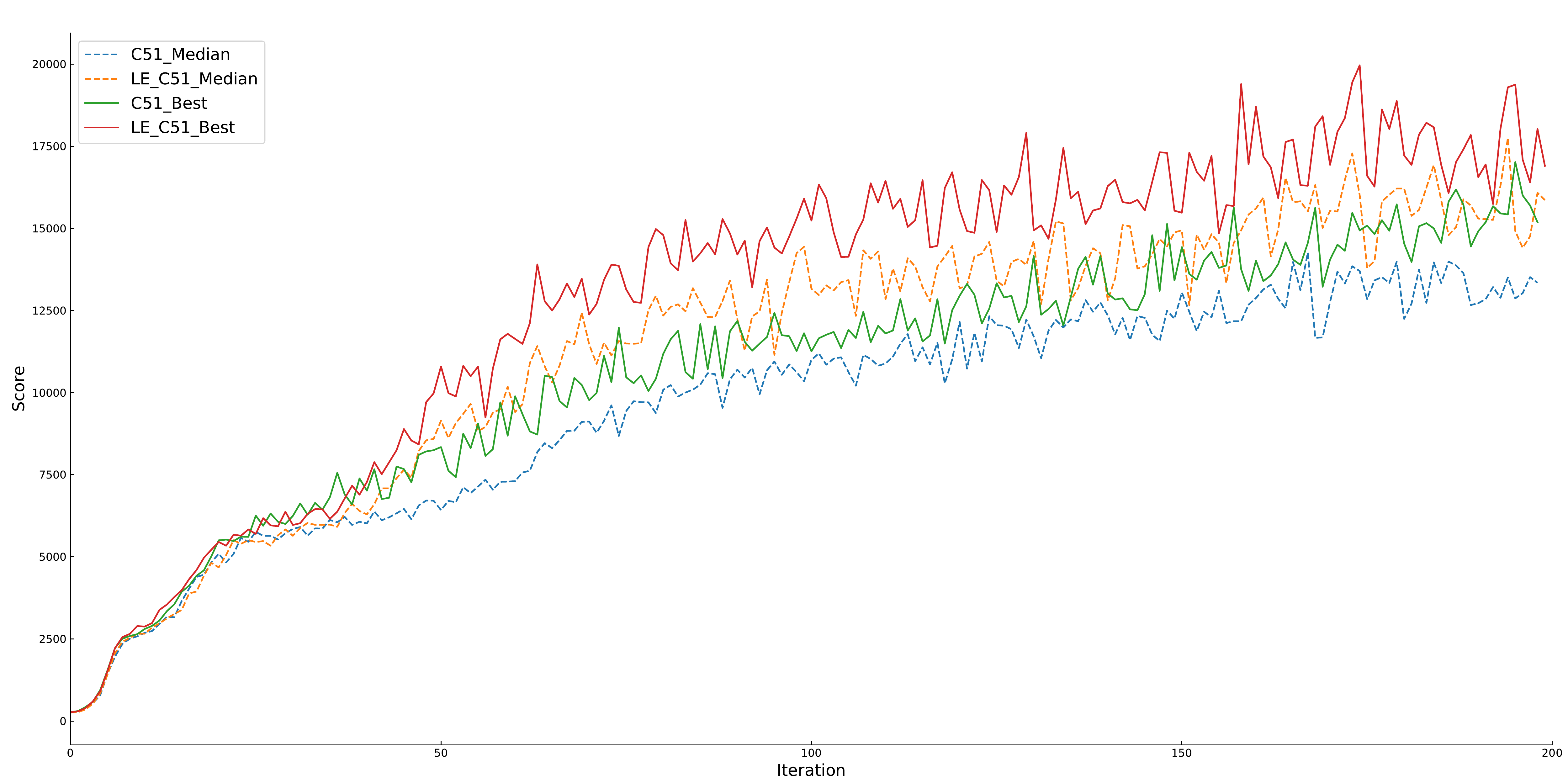}
\end{minipage}%
}%

\subfigure[LE\_IQN]{
\begin{minipage}[b]{0.23\textwidth}
\centering
\includegraphics[width=1\textwidth]{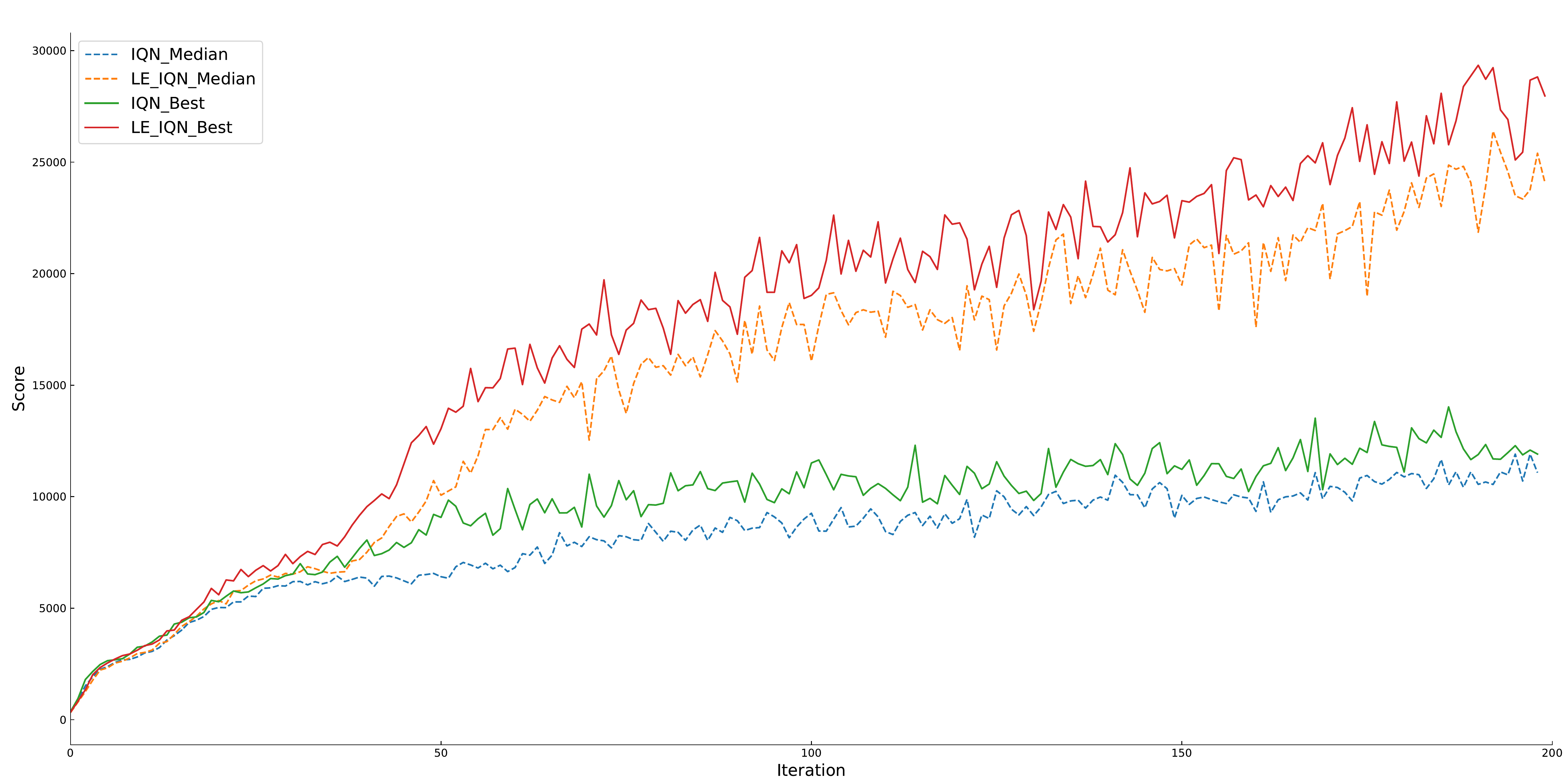}
\end{minipage}
}%
\subfigure[LE\_Rainbow(10)]{
\begin{minipage}[b]{0.23\textwidth}
\centering
\includegraphics[width=1\textwidth]{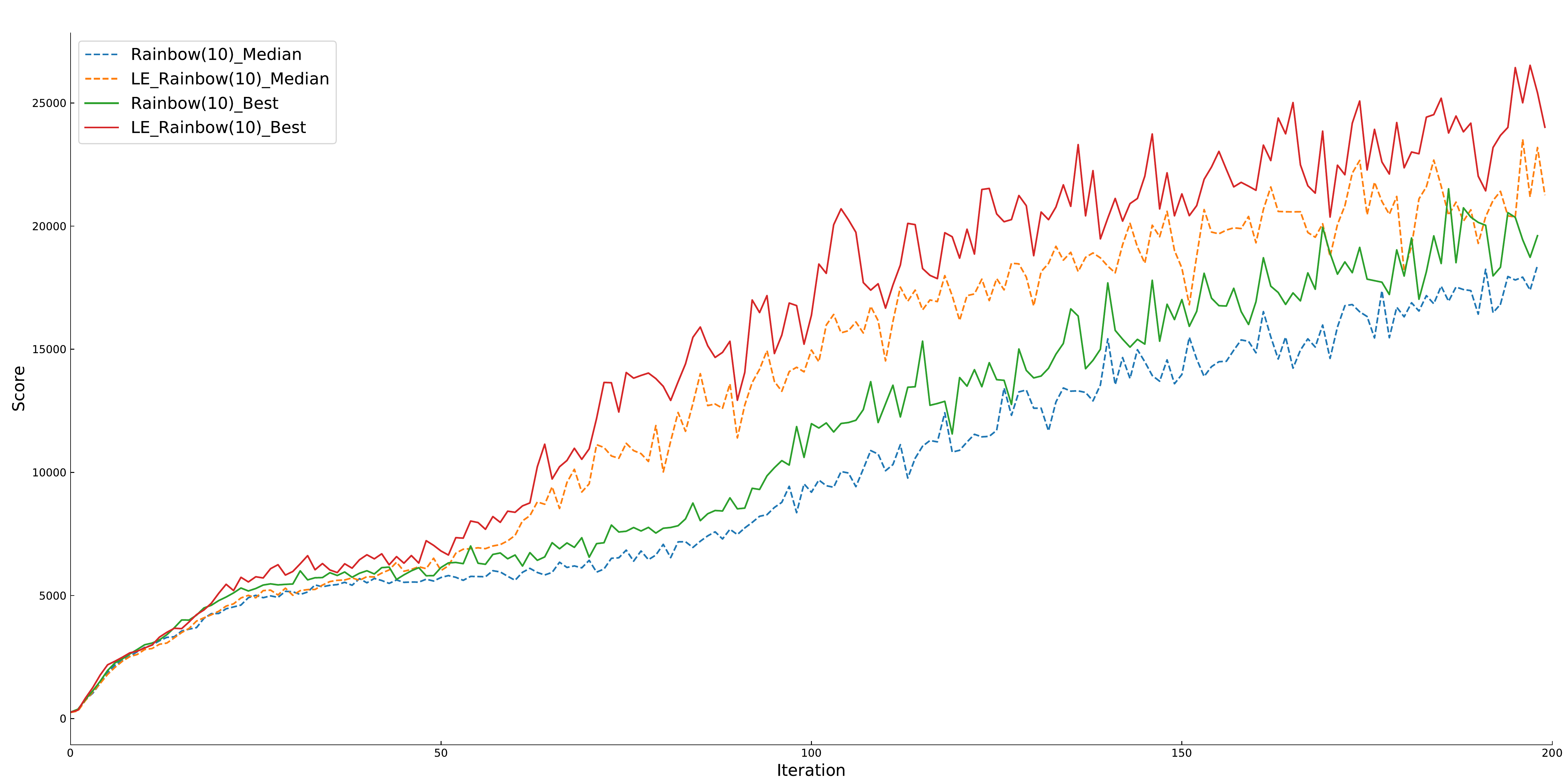}
\end{minipage}
}%

\centering
\caption{The Best Performance of Different LERL Algorithms in Asterix}
\end{figure}

In addition to the experiment of Rainbow algorithm, we also test the LERL algorithm using DQN, C51 or IQN in Asterix. The default evolution cycle of related experiments is still 5 iterations. (a-c) in Figure 4 show the best and median performance of LE\_DQN, LE\_C51 and LE\_IQN in Asterix. We also try to adjust the evolution cycle and give the performance data of LE\_Rainbow with the evolution cycle of 10 iterations, as shown in Figure 4(d). As can be seen from Figure 4, in addition to LE\_DQN, the other two kinds of LERL algorithms have achieved a large degree of performance improvement. Combining with the performance data of LE\_Rainbow (10) in Asterix, it can be analyzed that in Asterix environment, the expression ability of neural network is the main limiting factor for the performance improvement of LERL (function bottleneck).

\section{CONCLUSION}

We propose a general agent population learning system (GAPLS) and give the corresponding algorithm flow according to its system structure. On this basis, we propose lineage evolutionary reinforcement learning algorithm (LERL) which combines genetic algorithm, reinforcement learning algorithm and lineage factor. The performance of LERL in the experiment reached our overdue and verified our theoretical analysis. Compared with the original reinforcement learning algorithm, the performance and learning speed of the algorithm with LE are improved. In the future research, we think that some improvements in mutation operations and crossover operations may be able to reduce the performance fluctuations caused by evolution operations. In addition, the initial values of some parameters in evolution operations also can be further studied.

\bibliographystyle{plain}
\bibliography{uai20}

\begin{thebibliography}{10}

\bibitem{barth2018distributed}
Gabriel Barth-Maron, Matthew~W Hoffman, David Budden, Will Dabney, Dan Horgan,
  Dhruva Tb, Alistair Muldal, Nicolas Heess, and Timothy Lillicrap.
\newblock Distributed distributional deterministic policy gradients.
\newblock {\em arXiv preprint arXiv:1804.08617}, 2018.

\bibitem{bellemare2017distributional}
Marc~G Bellemare, Will Dabney, and R{\'e}mi Munos.
\newblock A distributional perspective on reinforcement learning.
\newblock In {\em Proceedings of the 34th International Conference on Machine
  Learning-Volume 70}, pages 449--458. JMLR. org, 2017.

\bibitem{bellemare2013arcade}
Marc~G Bellemare, Yavar Naddaf, Joel Veness, and Michael Bowling.
\newblock The arcade learning environment: An evaluation platform for general
  agents.
\newblock {\em Journal of Artificial Intelligence Research}, 47:253--279, 2013.

\bibitem{castro2018dopamine}
Pablo~Samuel Castro, Subhodeep Moitra, Carles Gelada, Saurabh Kumar, and Marc~G
  Bellemare.
\newblock Dopamine: A research framework for deep reinforcement learning.
\newblock {\em arXiv preprint arXiv:1812.06110}, 2018.

\bibitem{dabney2018implicit}
Will Dabney, Georg Ostrovski, David Silver, and R{\'e}mi Munos.
\newblock Implicit quantile networks for distributional reinforcement learning.
\newblock {\em arXiv preprint arXiv:1806.06923}, 2018.

\bibitem{dabney2018distributional}
Will Dabney, Mark Rowland, Marc~G Bellemare, and R{\'e}mi Munos.
\newblock Distributional reinforcement learning with quantile regression.
\newblock In {\em Thirty-Second AAAI Conference on Artificial Intelligence},
  2018.

\bibitem{fortunato2017noisy}
Meire Fortunato, Mohammad~Gheshlaghi Azar, Bilal Piot, Jacob Menick, Ian
  Osband, Alex Graves, Vlad Mnih, Remi Munos, Demis Hassabis, Olivier Pietquin,
  et~al.
\newblock Noisy networks for exploration.
\newblock {\em arXiv preprint arXiv:1706.10295}, 2017.

\bibitem{gazzaniga2000new}
Michael~S Gazzaniga.
\newblock {\em The new cognitive neurosciences}.
\newblock MIT press, 2000.

\bibitem{gu2016continuous}
Shixiang Gu, Timothy Lillicrap, Ilya Sutskever, and Sergey Levine.
\newblock Continuous deep q-learning with model-based acceleration.
\newblock In {\em International Conference on Machine Learning}, pages
  2829--2838, 2016.

\bibitem{hansen2006cma}
Nikolaus Hansen.
\newblock The cma evolution strategy: a comparing review.
\newblock In {\em Towards a new evolutionary computation}, pages 75--102.
  Springer, 2006.

\bibitem{hessel2018rainbow}
Matteo Hessel, Joseph Modayil, Hado Van~Hasselt, Tom Schaul, Georg Ostrovski,
  Will Dabney, Dan Horgan, Bilal Piot, Mohammad Azar, and David Silver.
\newblock Rainbow: Combining improvements in deep reinforcement learning.
\newblock In {\em Thirty-Second AAAI Conference on Artificial Intelligence},
  2018.

\bibitem{mnih2016asynchronous}
Volodymyr Mnih, Adria~Puigdomenech Badia, Mehdi Mirza, Alex Graves, Timothy
  Lillicrap, Tim Harley, David Silver, and Koray Kavukcuoglu.
\newblock Asynchronous methods for deep reinforcement learning.
\newblock In {\em International conference on machine learning}, pages
  1928--1937, 2016.

\bibitem{mnih2013playing}
Volodymyr Mnih, Koray Kavukcuoglu, David Silver, Alex Graves, Ioannis
  Antonoglou, Daan Wierstra, and Martin Riedmiller.
\newblock Playing atari with deep reinforcement learning.
\newblock {\em arXiv preprint arXiv:1312.5602}, 2013.

\bibitem{mnih2015human}
Volodymyr Mnih, Koray Kavukcuoglu, David Silver, Andrei~A Rusu, Joel Veness,
  Marc~G Bellemare, Alex Graves, Martin Riedmiller, Andreas~K Fidjeland, Georg
  Ostrovski, et~al.
\newblock Human-level control through deep reinforcement learning.
\newblock {\em Nature}, 518(7540):529--533, 2015.

\bibitem{schaul2015prioritized}
Tom Schaul, John Quan, Ioannis Antonoglou, and David Silver.
\newblock Prioritized experience replay.
\newblock {\em arXiv preprint arXiv:1511.05952}, 2015.

\bibitem{sutton1998introduction}
Richard~S Sutton, Andrew~G Barto, et~al.
\newblock {\em Introduction to reinforcement learning}, volume 135.
\newblock MIT press Cambridge, 1998.

\bibitem{van2016deep}
Hado Van~Hasselt, Arthur Guez, and David Silver.
\newblock Deep reinforcement learning with double q-learning.
\newblock In {\em Thirtieth AAAI conference on artificial intelligence}, 2016.

\bibitem{wang2015dueling}
Ziyu Wang, Tom Schaul, Matteo Hessel, Hado Van~Hasselt, Marc Lanctot, and Nando
  De~Freitas.
\newblock Dueling network architectures for deep reinforcement learning.
\newblock {\em arXiv preprint arXiv:1511.06581}, 2015.

\end{thebibliography}
\cite{dabney2018implicit,barth2018distributed,hansen2006cma,sutton1998introduction,
castro2018dopamine,bellemare2013arcade,bellemare2017distributional,
gazzaniga2000new,mnih2013playing,mnih2016asynchronous,gu2016continuous,
mnih2015human,van2016deep,schaul2015prioritized,wang2015dueling,
dabney2018distributional,hessel2018rainbow,fortunato2017noisy}

\end{document}